\documentclass[10pt,twocolumn]{article}
\usepackage{mwe}
\usepackage{cvpr}
\usepackage{times}

\cvprfinalcopy 

\usepackage{graphicx}
\usepackage{mwe}
\usepackage{epsfig}
\usepackage{graphicx}
\usepackage{amsmath}
\usepackage{amssymb}
\usepackage{enumitem}
\usepackage[pagebackref=true,breaklinks=true,letterpaper=true,colorlinks,bookmarks=false]{hyperref}
\usepackage[utf8]{inputenc} 
\usepackage[T1]{fontenc}    
\usepackage{hyperref}       
\usepackage{booktabs}       
\usepackage{amsfonts}       
\usepackage{nicefrac}       
\usepackage{microtype}      
\usepackage[algoruled
,boxed,lined]{algorithm2e}
\usepackage{graphicx}
\usepackage{arydshln}
\usepackage{titling}

\usepackage[export]{adjustbox}
\usepackage{xcolor}
\definecolor{blue}{rgb}{0,0.,0.}
\usepackage{colortbl}
\usepackage{array,colortbl,ctable}
 \usepackage{array,multirow}
\usepackage{subfig}
\newcommand{\suttl}[0]{MimicGAN}
\usepackage{siunitx}
\usepackage{color}
\usepackage{multirow}
\usepackage{authblk}
\begin{document}
\setlength{\droptitle}{-5em}   

\title{\textbf{MimicGAN: Robust Projection onto Image Manifolds with Corruption Mimicking}}


 \author[]{Rushil Anirudh\thanks{Corresponding author: \href{mailto:anirudh1@llnl.gov}{anirudh1@llnl.gov}}, ~Jayaraman J. Thiagarajan, ~Bhavya Kailkhura, ~Timo Bremer}
 \affil[]{Center for Applied Scientific Computing (CASC), \\ Lawrence Livermore National Laboratory}


\date{}

\maketitle

\begin{abstract} 
In the past few years, Generative Adversarial Networks (GANs) have dramatically advanced our ability to represent and parameterize high-dimensional, non-linear image manifolds. As a result, they have been widely adopted across a variety of applications, ranging from challenging inverse problems like image completion, to problems such as anomaly detection and adversarial defense. A recurring theme in many of these applications is the notion of projecting an image observation onto the manifold that is inferred by the generator. In this context, Projected Gradient Descent (PGD) has been the most popular approach, which essentially optimizes for a latent vector that minimizes the discrepancy between a generated image and the given observation. However, PGD is a brittle optimization technique that fails to identify the right projection (or latent vector) when the observation is corrupted, or perturbed even by a small amount. Such corruptions are common in the real world, for example images in the wild come with unknown crops, rotations, missing pixels, or other kinds of non-linear distributional shifts which break current encoding methods, rendering downstream applications unusable. To address this, we propose corruption mimicking -- a new robust projection technique, that utilizes a surrogate network to approximate the unknown corruption directly at test time, without the need for additional supervision or data augmentation. The proposed method is significantly more robust than PGD and other competing methods under a wide variety of corruptions, thereby enabling a more effective use of GANs in real-world applications. More importantly, we show that our approach produces state-of-the-art performance in several GAN-based applications -- anomaly detection, domain adaptation, and adversarial defense, that benefit from an accurate projection.

\let\thefootnote\relax \footnotetext{This work was performed under the auspices of the U.S. Department of Energy by Lawrence Livermore National Laboratory under Contract DE-AC52-07NA27344.}
\end{abstract}

\section{Introduction}
Generative Adversarial Networks (GANs)~\cite{GANGoodfellow} have been widely adopted in part due to their ability to parameterize complex, high-dimensional, non-linear image manifolds. As a result, they have been effective in several applications ranging from super resolution \cite{ledig2016photo} and image editing~\cite{pathak2016context,yeh2017semantic}, to image-to-image translation \cite{zhu2017unpaired,liu2017unsupervised}, etc. Additionally, an accurate parameterization of the image manifold provides a powerful regularization -- sometimes referred as a generative prior -- for a range of problems such as defense against adversarial attacks~\cite{defenseGAN,ilyas2017robust}, anomaly detection \cite{akcay2018ganomaly,zenati2018efficient}, and compressive sensing \cite{shahICASSP2018}. Despite the variabilities in their goals and formulations, an overarching requirement in all these applications is the ability to project an image observation onto the image manifold at test time.
\vspace{5pt}

\begin{figure*}[!htb]
	\centering

	\includegraphics[trim={1.0cm 7.5cm 1.0cm 0.0cm},clip,width=0.85\linewidth,valign=c]{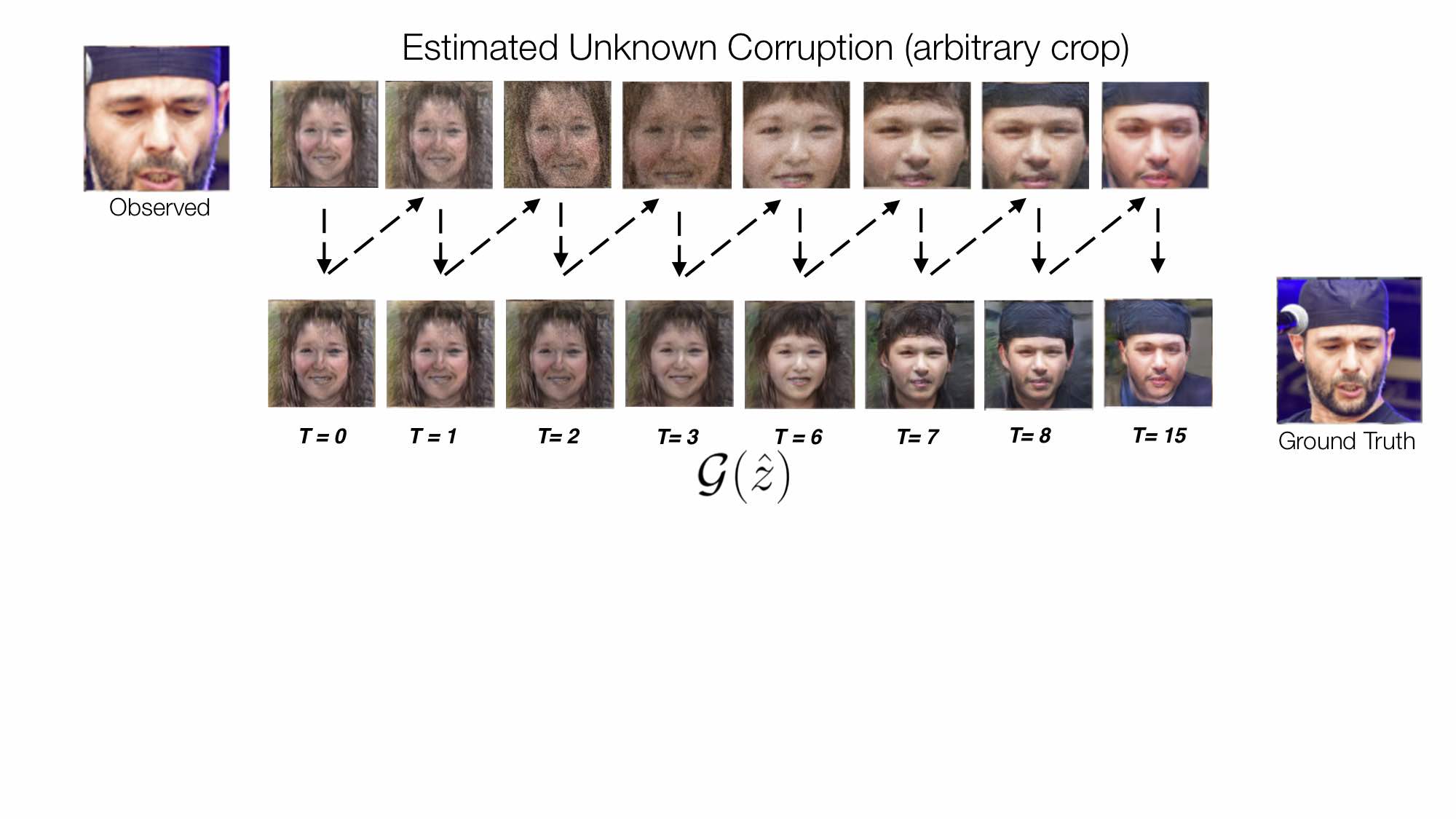}
		
	\caption{Demonstration of the alternating optimization process. Though initialized randomly, the corruption-mimicking network effectively guides the search in the latent space of the generator to produce an estimate of the true image. In this example from the FFHQ dataset \cite{karras2018style}, the observed image is a cropped (\textit{zoom in}) version of the ground truth.}
	
	\label{fig:alternating}
\end{figure*}

\paragraph{} Projecting onto an image manifold essentially involves an \textit{encoding} step to project onto the latent space of a generator, followed by a \textit{decoding} step to obtain the image from the image manifold. While the latter is carried out using a pre-trained generator model (e.g. convolutional neural networks or CNNs), the former step is typically implemented using two distinct family of approaches: (a) Projected gradient descent (PGD): a simple, yet powerful, latent space optimization strategy that minimizes the discrepancy between an observed sample, and the best estimate of that sample from the generator~\cite{yeh2017semantic,defenseGAN}; and (b) Coupling GAN with an explicit encoder: Techniques such as, BiGAN~\cite{donahue2016adversarial} and ALI~\cite{dumoulin2016ALI}, include an additional network that explicitly provides a latent space representation for an image. This broad class of approaches also encompasses other generative modeling techniques such as variational autoencoders (VAEs) \cite{kingma2013auto}, \cite{larsen2015autoencoding,bojanowski2017optimizing}.

\noindent Though these techniques have been shown to provide effective projections, they are easily broken and fail when the sample to be projected has undergone even minor distributional shifts compared to the training data, for e.g. shifts like simple corruptions, missing pixels, translations, rotation or scale changes etc. Inferencing in the latent space of GANs has gained a lot of attention recently \cite{GradientPub,abdal2019,encoderStyleGAN} with the advent of high-quality GANs such as BigGAN~\cite{brock2018large}, and StyleGAN~\cite{karras2018style}, thus strengthening the need for robust projections. While existing solutions such as unsupervised image-to-image translation \cite{zhu2017unpaired,liu2017unsupervised} have been remarkably successful in modeling complex distributional shifts across image domains, they rely on the availability of a large set of samples from the two domains. In contrast, we are interested in robustness to simpler distribution shifts such as affine transformations, corruptions or missing information, that necessitate robust projections even when very few samples are available from the corrupted domain at test time.

\paragraph{} In this paper, we propose \suttl, a test-time approach for projecting images into the latent space of pre-trained generative models like GANs such that the decoder remains robust to a wide range of distribution shifts. Our main contribution is the process of \emph{corruption-mimicking} which uses a surrogate neural network to mimic the corruption process, while simultaneously computing the projection. Given a small number of observations, we perform an alternating optimization to train the surrogate conditioned on the current best estimate of projections from the generator, followed by optimization in the latent space to identify the best projections conditioned on the current estimate of the surrogate network. Since the surrogate is a shallow network, and a generative prior acts as a powerful regularizer, robust projection can be achieved with as few as 2 samples in some cases, or as many as required by the complexity of the application. Importantly, because we estimate the corruption on the fly, \suttl~\emph{does not need to know, in advance, how the images have been corrupted}, or any prior assumptions on the type of distribution shift that has occurred. The alternating optimization procedure of \suttl~is demonstrated in Figure \ref{fig:alternating}.


\paragraph{\textbf{Main Findings}}
We observe that corruption mimicking leads to projections that are highly robust to changes in scale, translation, and rotation; to missing or partial information, and to domain shifts across similar datasets. It needs to be emphasized that the robustness is not achieved by using any kind of data augmentation strategies, but rather as a consequence of the corruption-mimicking, while keeping all other conditions the same. Broadly, \suttl~ demonstrates robustness to the class of functions that can be expressed by the surrogate network considered. Moreover, PGD is a special case of \suttl~ when the surrogate network is assumed to be the identity function and this simplification leads to highly sub-optimal projections in all applications considered.

\vspace{5pt}

\paragraph{\textbf{Contributions}}

\begin{itemize}[leftmargin=*,itemsep=.1mm]
  \item[1.] We propose a robust version of the popularly adopted PGD approach and its variants called \suttl, that achieves robustness across a wide variety of test-time corruptions such as scaling, rotation, and missing or partial information.
  \item[2.] We propose a corruption mimicking surrogate network that estimates unknown distributional shifts or corruptions on the fly, resulting in a single algorithm that works across a wide range of test-time distribution shifts, without any prior knowledge.
  \item[3.] Using comprehensive experimental evaluation, we show that \suttl~is significantly more robust in terms of projection error compared to several competitive encoding techniques trained on the same data, with no additional supervision.
  \item[4.] We demonstrate significant performance improvements with \suttl~in applications benefitting from accurate projections: adversarial defense, anomaly detection, domain adaptation, and adapting GANs across datasets (e.g. MNIST\cite{lecun1998mnist}, USPS hand written digits, CelebA \cite{liu2015faceattributes}, FFHQ-Thumbnails \cite{karras2018style}, LFW \cite{becker2013evaluating}). 
\end{itemize}

\section{Related Work}
\suttl~ bridges the gap between two diverse yet related GAN-based methods in literature. On the one hand, a growing number of applications rely on using the image manifold as a powerful regularizer for inverse problems, adversarial defense, anomaly detection etc. On the other hand, GAN-style training has emerged as a powerful tool for image-to-image translation, that are able to map across very complex distributional shifts. \suttl~enjoys the merits of both approaches, wherein it is highly effective at leveraging the image manifold for adversarial defense, while also modeling the image translation problem as computing projections onto the manifold. In this section, we will briefly review the current art in the two broad classes of approaches and present comparisons to the proposed \suttl.

\paragraph{\textbf{Image-to-Image Translation}} While several recent techniques have been proposed to handle very complex distributional shifts, such as unpaired image to image translations \cite{zhu2017unpaired,liu2017unsupervised}, Pix2Pix\cite{isola2017image} etc.,  they involve training networks that are able to map from one distribution to the other. More recently, Non Adversarial Mappings (NAMs) \cite{hoshen2018non,hoshen2018nam} have shown preliminary evidence to find translations across domains without adversarial training. NAMs are conceptually similar to our approach, since they parameterize the distribution shift by a neural network that relies on a pre-trained generator. However, their approach expects a large set of observations (over $2000$, \cite{hoshen2018non}) for a type of corruption, and as a result needs to be re-trained each time. Further, none of the existing image-to-image translation techniques effectively leverages a GAN for projection, which is the primary focus of this work. For example, it is not clear how they could be used for applications, such as, adversarial defense or anomaly detection, because of their large observed data requirement. Our goal is to perform robust projection at the test time, that can work with as few as a 5-10 observations. Consequently, \suttl~ is able to project robustly across several distribution shifts using the same system and hyper-parameters with no additional tuning across the distribution shifts. 

\paragraph{\textbf{The image manifold as a prior:}} Our work improves on the notion of \textit{GAN priors} \cite{yeh2017semantic,shahICASSP2018}, \cite{AsimDeblurGAN}, \cite{bora2017compressed} -- the idea that optimizing in the latent space of a pre-trained GAN provides a powerful prior to solve several traditionally hard problems. Yeh \textit{et al.}~\cite{yeh2017semantic} first introduced the projected gradient descent (PGD) method with GANs for filling up arbitrary holes in an image, in a semantically meaningful manner. Subsequently, several efforts have pursued solving inverse problems using PGD, for example compressive recovery~\cite{bora2017compressed}, \cite{shahICASSP2018}, and deblurring~\cite{AsimDeblurGAN}. Asim \textit{et al.}~\cite{AsimDeblurGAN} proposed an alternative approach for image deblurring, wherein they used two separate GANs -- one for the blur kernel, and another for the images. \suttl~differs from this approach in that we make no assumptions on the types of corruptions that can be recovered, and hence requires only a single GAN for modeling the image manifold. There have also been improvements to the recovery of latent vectors for GANs \cite{lipton2017precise}, where it is shown that stochastic clipping can provide more accurate projections. This technique works effectively on clean data, but suffers from all the problems of PGD, in being non-robust. At the same time, it is a generic approach which can be combined with \suttl for improved performance. {\color{blue} Finally, \suttl~ is also related to iGAN \cite{zhu2016generative}, where the projection on the image manifold is used for user-guided image manipulation. The projection is achieved by an encoder that is trained at test time to find the right initialization in the GAN latent space. We compare with iGAN in our experiments, and show that while it is more robust than PGD, it is susceptible to many of its weaknesses.}

\vspace{5pt}
\begin{figure*}[!htb]
	\centering
	{\includegraphics[trim={0.0cm 0.0cm 0.5cm 0.5cm},clip,width=0.75\linewidth]{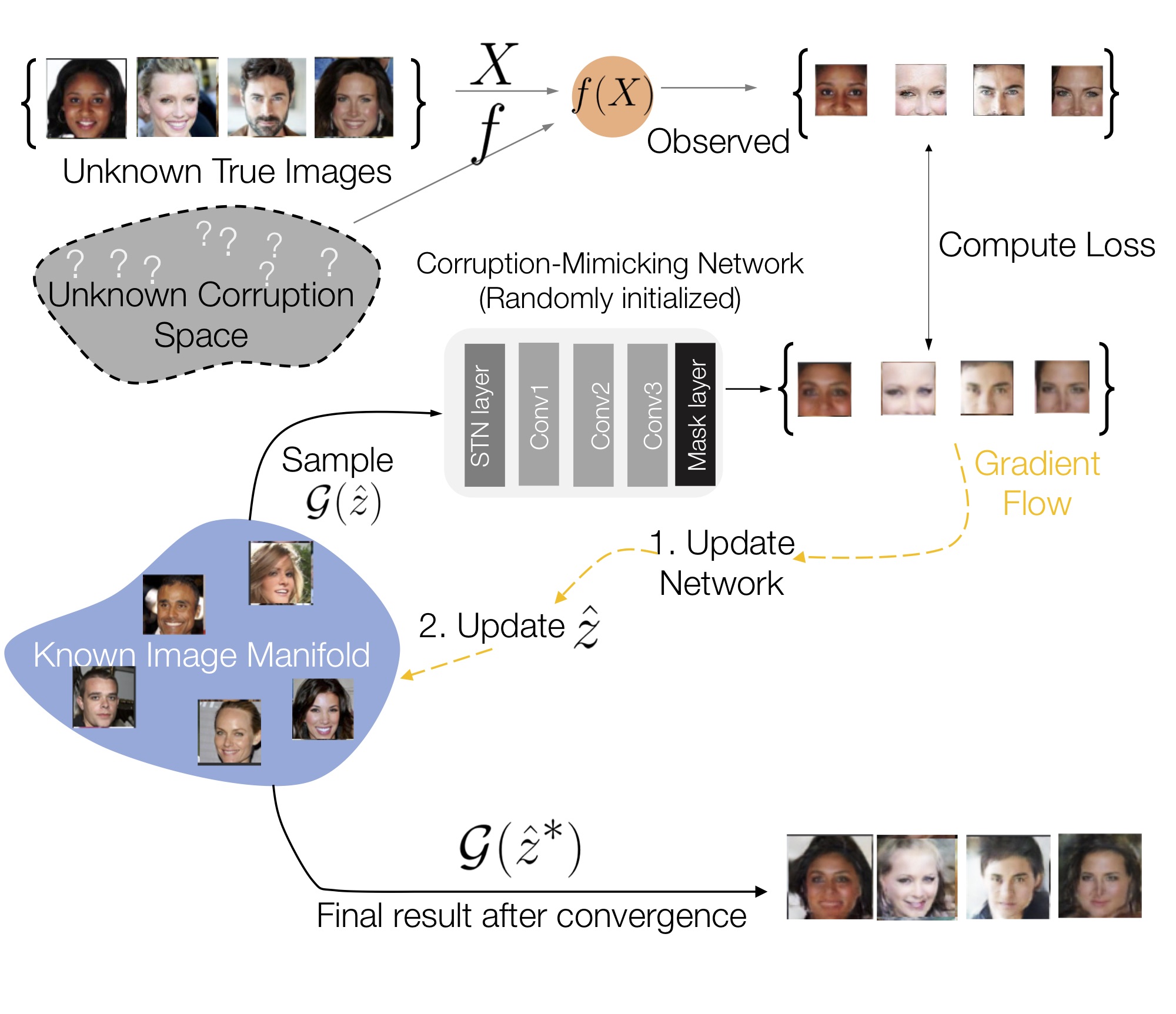}}	
	\caption{\suttl: Illustration of the overall approach for projecting test images onto a known image manifold parameterized using a GAN.}
	\label{fig:mimic}
\end{figure*} 

\noindent  In addition to its effectiveness for image recovery, the GAN-prior has also become a reasonably successful way to defend against adversarial attacks -- which are small perturbations in the pixel space that are designed to cause a particular classifier to fail dramatically. For example, Defense-GAN \cite{defenseGAN}, uses an approach called \textit{clean and classify}, where the observed adversarial example is projected on to the image manifold, with the hope of eliminating the adversarial perturbation in the process. A very related idea is explored in \cite{ilyas2017robust}, referred as \textit{Invert and Classify} (INC), which also relies on PGD similar to Defense-GAN. A similar approach proposes to use the discriminator in addition to the generator to detect adversarial examples \cite{santhanam2018defending}, and we show in our experiments that, in comparison, the \suttl defense is significantly more robust. Interestingly, all the three aforementioned approaches become a special case of {\suttl}, when the surrogate network in our approach is assumed to be identity. Note that, both Defense-GAN and {\suttl~} are applicable to white-box as well as black-box attacks, since they do not need access to the classifier model in order to clean the data. It should be noted that most existing GAN based defenses are effective only when the attack is designed purely on the classifier. However, recent evidence shows that these defenses can be broken when the attack is designed on the GAN and the classifier together \cite{athalye2018obfuscated}. We test \suttl~ against such an attack and observe that, it is also vulnerable to such a GAN-based attack, but to a lower degree than existing GAN-based defense strategies.

{\color{blue}
\paragraph{\textbf{Structural network priors}} Finally, our work is also related to the category of recent approaches that leverage the prior afforded by an untrained neural network itself to solve traditionally challenging inverse problems. The use of an untrained surrogate network to estimate the corruption draws inspiration from previous methods such as Deep Image Prior \cite{ulyanov2017deep} and Deep Internal Learning \cite{shocher2018zero}, which work directly at test time. In the case of \suttl, the surrogate network architecture imposes a structural prior on the types of corruptions that can be recovered accurately, as described in section \ref{sec:architecture}.}
\vspace{-10pt}
\section{Proposed Approach}
\label{sec:proposed}
In this section, we describe \suttl~in detail, wherein the core idea is to jointly obtain an estimate of the unknown corruption process and perform projection via latent space optimization. The proposed algorithm is generic in that it can work with any type of generative model including the families of GANs and variational autoencoders, as we make no assumptions on the generator except that we are able to sample from a latent space in a differentiable manner.

\subsection{\textbf{Formulation: Corruption-Mimicking for Robust Projection}}
Let us denote the generator of a pre-trained GAN as $\mathcal{G}:\mathrm{R}^d\mapsto \mathcal{I}$, where $\mathcal{I}$ represents the image manifold and $d$ is the dimensionality of the latent space. Assuming the observation that needs to be projected onto $\mathcal{I}$ is given by $Y^{obs} = f(X)$, where $f(.)$ is some unknown corruption function, and $X \in \mathcal{I}$ is the true image on the manifold. Note that, {\color{blue} $f$ belongs to a broad class of functions} including geometric transformations (e.g. affine shift), missing information (e.g. missing pixels), or a systematic distributional shift (e.g. changes to pixel intensities). The goal of robust projection is to estimate $X$ by projecting $Y^{obs}$ onto $\mathcal{I}$ for a wide class of functions or corruptions, without any prior knowledge about $f$ or paired examples representing the function. Formally, let $\mathbf{Y}^{obs} = \{Y^{obs}_1,Y^{obs}_2,\dots,Y^{obs}_N\}$, a small number of observations such that $N>1$. These are assumed to be produced by a corruption process $f$ along with an unknown noise component, i.e., ${Y}^{obs}_j = f\left({X}_j, \Theta_f\right)$, where $X_j$'s are assumed to be i.i.d, drawn at random from the image manifold parameterized by the generator $\mathcal{G}$; and $\Theta_f \sim \mathcal{N}(\mathbf{\mu}_f,\mathbf{\sigma}_f)$ correspond to the unknown corruption parameters, assumed to be drawn from a Gaussian with mean and variance denoted by $\mathbf{\mu}_f$ and $\mathbf{\sigma}_f$ respectively.

For e.g., $\mathbf{\mu}_f$ could represent the parameters of an affine transform, or even represent a non-linear distribution shift (such as across datasets). In this formulation $f$ can be inherently stochastic, i.e., different observations can be corrupted in slightly different ways. In the rest of the text we drop the notation $\Theta_f$ for convenience.
\vspace{5pt}

Our main contribution in this paper is a technique called \emph{corruption mimicking}, that is able to estimate the likely corruption function while simultaneously projecting the observation onto the image manifold. We propose to parameterize the corruption function by a neural network, $\hat{f}$, that ``mimics'' the unknown corruption process, conditioned upon the current best estimate of the projection $\hat{X}\in \mathcal{I}$. Next, we estimate the best possible projection $\hat{X}$ conditioned on the current estimate of the function $\hat{f}$. This alternating optimization progresses by incrementally refining $\hat{f}$, while ensuring that there always exists a valid solution $\hat{X} \in \mathcal{I}$, such that $\hat{f}(\hat{X}) \approx Y^{obs}$. The generator is kept frozen during the entire optimization loop, and hence this is implemented as an entirely test-time solution. Figure \ref{fig:mimic} shows an overview of the proposed approach, and how the gradients are used to update $\hat{f}$, and $z$ respectively. Figure \ref{fig:alternating} shows an example of the alternating optimization procedure, that demonstrates how a better estimate of $\hat{f}$ can lead to a reliable projection onto $\mathcal{I}$. In both the examples shown, the corruption function is arbitrary cropping (``zoom in'') applied to the original image, which the corruption surrogate, $\hat{f}$, captures within very few iterations. 

\noindent The proposed solution is applicable over a wide class of functions $f \in \mathcal{F}$, where $\mathcal{F}$ is the hypothesis space of corruption functions that can be modeled using the surrogate neural network. For notational convenience, we use $\hat{f}$ to represent the corruption surrogate network, and $\hat{\Theta}$ to represent its weights. Mathematically, we formulate corruption mimicking as follows:

\begin{equation}
\centering
\label{eq:obj}
\hat{\Theta}^*,\{z^*_j\}_{j=1}^N = \underset{\hat{\Theta},\{z_j\in \mathbb{R}^{d}\}_{j=1}^N}{\arg\min} \;
\sum_{j = 1}^N \mathcal{L}\left(Y_j^{obs}, \hat{f}(\mathcal{G} (z_j),\hat{\Theta})\right),
\end{equation} 
Since both $\mathcal{G}$ and $\hat{f}$ are differentiable, we can evaluate the gradients of the objective in \eqref{eq:obj}, using backpropagation and utilize existing gradient based optimizers. In addition to computing gradients with respect to $z_j$, we also perform clipping in order to restrict it within the desired range (e.g., $[-1,1]$) resulting in a PGD-style optimization. Solving this alternating optimization problem produces high quality estimates for both $\hat{f}$ and the projections $\hat{X}_j = \mathcal{G}(z^*_j)~\forall j$.

{\color{blue}In the special case where the corruption function $f$ is known \textit{a priori}, the projection problem can be simplified:
\begin{equation}
\centering
\label{eq:classic}
\{z^*_j\}_{j=1}^N = \underset{\{z_j \in \mathbb{R}^{d}\}_{j=1}^N}{\arg\min} \;
 \sum_{j = 1}^N \mathcal{L}\left(Y_j^{obs}, f(\mathcal{G}(z_j))\right),
\end{equation}
where the optimal projections are given by $\hat{X}_j = \mathcal{G}(z^*_j)$, $\mathcal{L}$ is a loss function (e.g. mean squared error) for reconstructing the observations using the known $f$ and $z \in \mathbb{R}^d$ is the latent noise vector. This is commonly adopted in several state-of-the-art solutions for image inpainting \cite{yeh2017semantic}, adversarial defense \cite{defenseGAN}, compressed sensing \cite{bora2017compressed},\cite{shahICASSP2018}, and deblurring \cite{AsimDeblurGAN}. 

\noindent Equation \eqref{eq:classic}, can be further simplified when we expect no corruption or distributional shift by setting $f = \mathbb{I}$, the identity transformation, which reduces the problem to one of plain projection. This is the most commonly used setting to obtain projections of clean images using generators, sometimes referred to as ``inversion'' of GANs \cite{creswell2018inverting,lipton2017precise}. In the remainder of the paper we refer to this setting as PGD.
\begin{equation}
\centering
\label{eq:pgd}
\{z^*_j\}_{j=1}^N = \underset{\{z_j \in \mathbb{R}^{d}\}_{j=1}^N}{\arg\min} \;
 \sum_{j = 1}^N \mathcal{L}\left(Y_j^{obs}, \mathcal{G}(z_j)\right),
\end{equation}

While \eqref{eq:classic} can be highly effective when $f$ is known, it fails dramatically when $f$ is unknown. In practice, this is often handled by making a na\"ive assumption that $f$ is identity as in \eqref{eq:pgd}, and employing PGD. This tends to produce poor results under a wide-range of commonly occurring corruptions, rendering the GAN essentially unusable in downstream applications.}

\subsubsection{Architecture of the corruption surrogate} 
\label{sec:architecture}
\begin{figure}[!htb]
	\centering
	\includegraphics[trim={0.0cm 0.0cm 0.0cm 0.0cm},clip,width=0.8\linewidth]{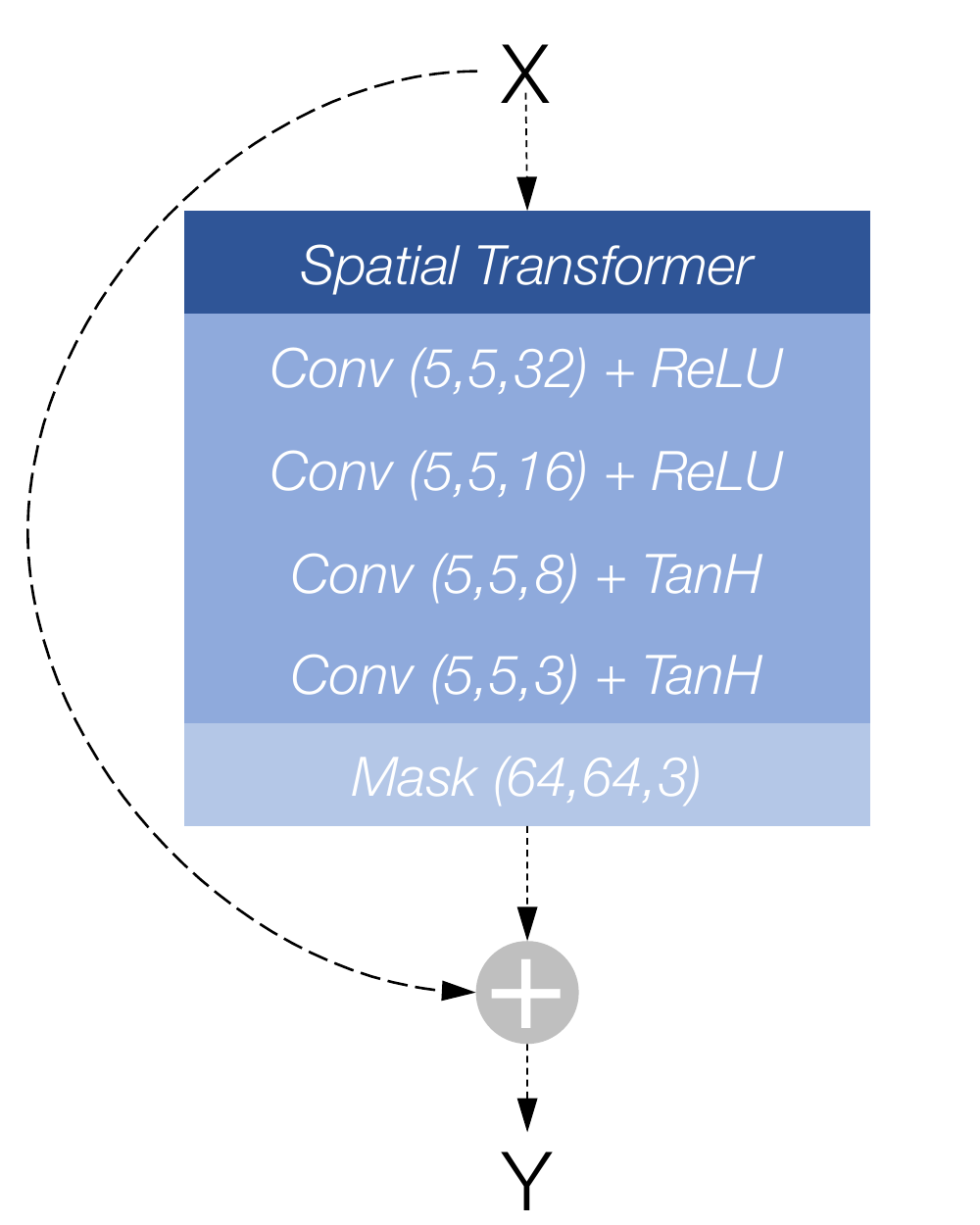}
	\caption{Architecture of surrogate network used to model the distribution shift at test time. The same architecture is used for CelebA, LFW and FFHQ datasets. {\color{blue}The mask acts independently on the RGB channels, as a multiplication without any bias. The addition symbol implies a passthrough that is just an addition operation, and we do not use any non-linearity after the sum.}}
	\label{fig:surrogate_Arch}
\end{figure}

The surrogate neural network plays a cruicial role in obtaining robust projections. We use a spatial transformer layer (STL) \cite{jaderberg2015spatial} first with the $6$ parameters that model affine transforms corresponding to scale, rotation and shearing. Next, we include $4$ convolutional layers, followed by a masking layer with the mask being randomly initialized with the same size of the image. The architecture is described in the Figure \ref{fig:surrogate_Arch}.

\paragraph{\textbf{Justification for the architecture}:} The robustness afforded by \suttl~corresponds to the class of functions that can be expressed using the surrogate network, $\hat{f}$. Consequently, we choose the layers based on commonly occuring corruptions. The spatial transformer captures affine transformations, as a result of which, we obtain robustness to geometric corruptions such as scale, rotation, shift. Next, convolutional layers with non-linearities allow us to model various image transformations such as blurs, edges, color shifts etc. The masking layer allows us to model corruptions such as arbitrary masks, missing pixels, noise, etc. Finally, we create a short-cut connection \cite{he2016deep} to make it easy for the network to model the identity transformation. As a result, \suttl~ can recover from corruptions corresponding to either the specific types or different combinations of them, for e.g. scaling and rotation, or scaling and masking etc.

\vspace{5pt}

\noindent The architecture of the surrogate network chosen in this work is based on ``typical'' corruptions we expect to see in the wild. The layers in the network can be substituted based on the application at hand, for example if one only expects to see affine corruptions then a ST layer should suffice for \suttl. More generally, by choosing the architecture for the surrogate network, we are imposing a \emph{structural prior} on the kinds of corruptions \suttl~ can handle. As a result, this can be a limitation if the corruption cannot be modeled by existing CNN layers, like affine transforms without the ST layer.
\vspace{5pt}

\noindent Given the highly underdetermined nature of this problem, in many cases $\hat{f}$ need not emulate the corruption exactly. Nevertheless, the process of learning this surrogate sufficiently regularizes the problem of computing the optimal projection. The balance between updating $\hat{f}$ and re-estimating the projection using PGD is critical to the convergence behavior. Since $\hat{f}$ is updated using a  different set of images in each iteration (due to updates in $z_j$), it rarely demonstrates overfitting behavior, even with as few as 2 observations. However, we note that, overfitting can occur if the surrogate network is made very deep, or there is only a single observation $(N=1)$. We investigate this behaviour further in section \ref{sec:properties}.

\subsubsection{Loss Function} 
Next we describe the construction of loss function $\mathcal{L}$ in (\ref{eq:obj}), which is comprised of two different terms:
\begin{itemize}

  \setlength\itemsep{-0.1em}
\item[(a)] \textbf{Corruption mimicking loss:} Measures the discrepancy between the true observations $Y_j^{obs}$ and the estimates from \suttl, $$\mathcal{L}_{obs} =  \sum_{j = 1}^N \left\|Y_j^{obs}-\hat{f}(\mathcal{G} ({z}_j))\right\|_1,$$ where $\|.\|_1$ is the $\ell_1$ norm. We observed that the $\ell_2$ norm works reasonably as well.
\item[(b)] \textbf{Adversarial loss:} Using the discriminator $\mathcal{D}$ and the generator $\mathcal{G}$ from the pre-trained GAN, we measure: $$\mathcal{L}_{adv} = \sum_{j=1}^N \log(1-\mathcal{D}(\mathcal{G}({z}_j)).$$ Note, this is the same as the generator loss used for training a GAN.
\end{itemize}Given these components, the overall loss is defined as:
\begin{equation}
  \label{eq:losses}
\hspace{50pt}  \mathcal{L} = \mathcal{L}_{obs} + \lambda_{adv}\mathcal{L}_{adv} ,
\end{equation}where $\lambda_{adv}$ is kept fixed at $1e-4$. Note that, the projected gradient descent (PGD) technique~\cite{yeh2017semantic,defenseGAN} can be derived as a special case of our approach, when ${f} = \mathbb{I}$, where $\mathbb{I}$ is identity. Also note that since the surrogate can never perfectly learn the identity function, \suttl~ will slightly underperform PGD in cases where there is no corruption, i.e. when the corruption is identity.

\begin{algorithm}[!htb]
\SetKwFunction{GAN}{GAN}
\SetKwFunction{random}{random}
\SetKwInOut{Input}{Input}
\SetKwInOut{Output}{Output}
\SetKwInOut{Initialize}{Initialize}
\Input{ Observed Images $\mathbf{Y}^{obs}$, Pre-trained generator $\mathcal{G}$ and discriminator $\mathcal{D}$.}
\Output{ Recovered Images $\hat{\mathbf{X}} \in \mathcal{I}$, Surrogate $\hat{f}$}
\Initialize{ For all $j, \hat{z}_j$ is initialized as average of $1000$ realizations drawn from $~~\mathcal{U}(-1,1)$\tcp{see text}}
\Initialize{Random initialization of surrogate parameters, $\hat{\Theta}$}

\BlankLine

\For{$t\leftarrow 1$ \KwTo $T$}{
    \BlankLine

  \For{$t_1\leftarrow 1$ \KwTo $T_1$ 
   \tcp{update surrogate}}{
    $Y_j^{est} \leftarrow \hat{f}\left(\mathcal{G}(\hat{z}_j); \hat{{\Theta}}\right),~\forall j$;

	Compute loss $\mathcal{L}$ using \eqref{eq:losses};

    $\hat{{\Theta}} \leftarrow \hat{{\Theta}} - \gamma_s~\nabla_{\hat{\Theta}}(\mathcal{L}) $;
    }

    \For{$t_2\leftarrow 1$ \KwTo $T_2$ 
    \tcp{perform PGD conditioned on surrogate}}{
      $Y_j^{est} \leftarrow \hat{f}\left(\mathcal{G}(\hat{z}_j);\hat{{\Theta}}\right),~ \forall j$;

	Compute loss $\mathcal{L}$ using \eqref{eq:losses};

      $\hat{z}_j \leftarrow \hat{z}_j - \gamma_g~\nabla_z(\mathcal{L}),~ \forall j$;

      $\hat{z}_j \leftarrow \mathcal{P}\left(\hat{z}_j\right)~ \forall j$;
      }
      \BlankLine

  }
  return $\hat{f}$, $\{\hat{{X}}_j = \mathcal{G}(\hat{z}_j),~\forall j$\}.
\caption\suttl\label{mainAlg}
\end{algorithm}

\subsubsection{Algorithm}
The procedure to perform the alternating optimization is shown in Algorithm \ref{mainAlg}. We run the inner loops for updating $\hat{f}$ and $z_j$ for $T_1$ and $T_2$ number of iterations respectively. The projection operation denoted by $\mathcal{P}$ is the clipping operation, where we restrict the $z_j$'s to lie in the range $[-1,1]$. We use the RMSProp Optimizer to perform the gradient descent step in each case, with learning rates of $\gamma_s$ and $\gamma_g$ for the two steps respectively. Note that, since our approach requires only the observations $\mathbf{Y}^{obs}$ to compute the projection, it lends itself to a task-agnostic inference wherein the user does not need to specify the type of corruption or acquire examples \emph{a priori}.

\subsubsection{Initialization}
\suttl~depends on an initial seed to begin the alternating optimization, and we observed large variabilities in convergence behavior due to the choice of the seed. In order to avoid this, we initialize the estimate of projected images by computing an average sample on the image manifold, by averaging $1000$ samples drawn from the random uniform distribution. When the latent space is drawn from a uniform distribution, this effectively initializes them with zero values. Note that, we initialize the estimate for all observations with the same mean image. We observe that this not only speeds up convergence, but is also stable across several random seeds. {\color{blue} There are more sophisticated initialization techniques such as in iGAN \cite{zhu2016generative}, however, these are effective under little or no corruptions, and fail severely otherwise. Zero-initialization also works just as effectively on the GANs considered here, but the proposed averaging technique is applicable more generally if latent space distributions are not zero-centered.}
\section{Robustness Experiments}
\label{sec:expt}
\begin{figure*}[!htb]
\centering
\subfloat[Robustness to rotations]{\includegraphics[trim={0.2cm 0.5cm 0cm 0.5cm},clip,width=0.48\linewidth]{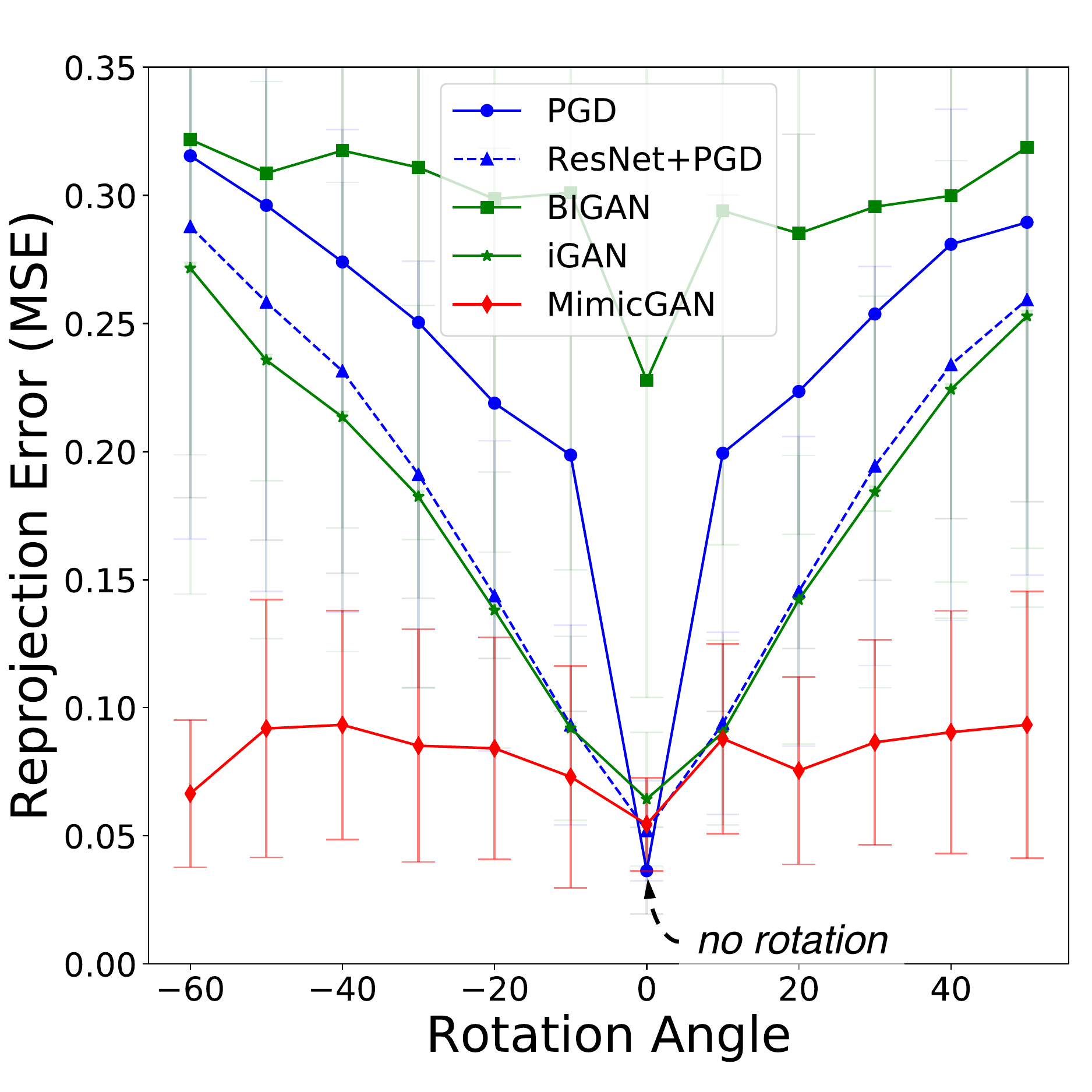}
\label{fig:rotation}}
\subfloat[Robustness to scale]{\includegraphics[trim={0.2cm 0.3cm 0cm 0.5cm},clip,width=0.48\linewidth]{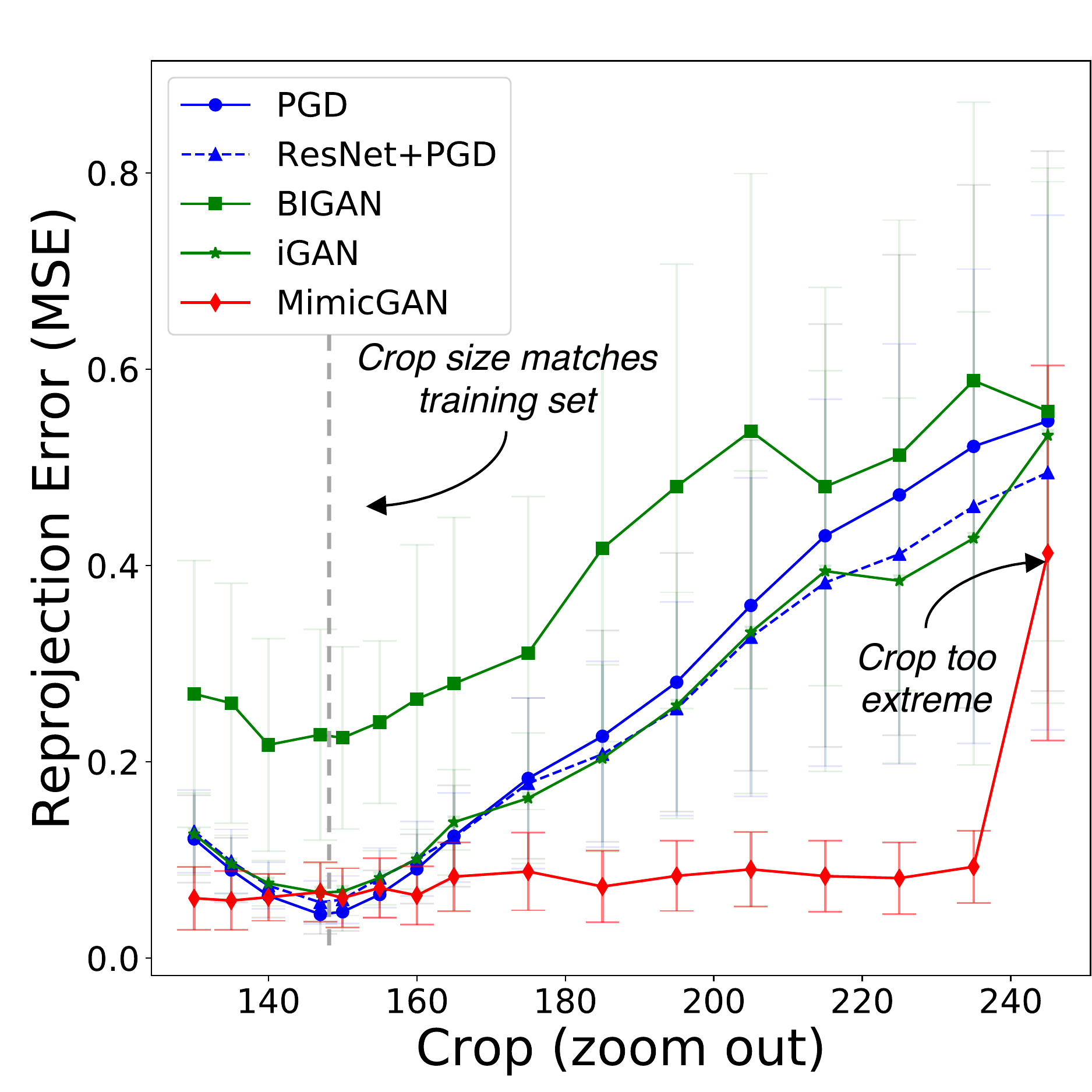}
\label{fig:crop}}

\subfloat[Projections under missing pixels]{\includegraphics[trim={0.2cm 0.5cm 0cm 0.5cm},clip,width=0.48\linewidth]{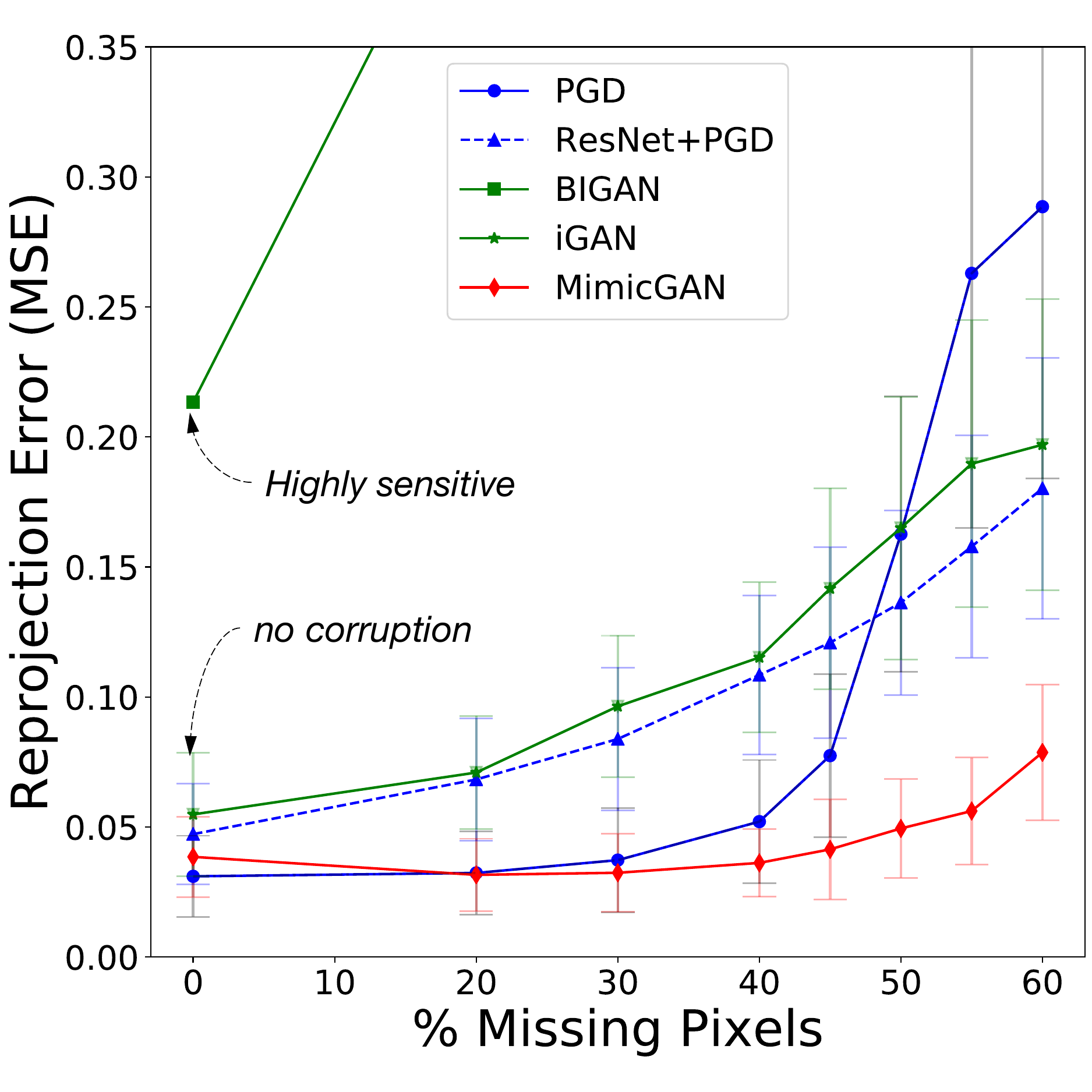}
\label{fig:missing}}
\subfloat[Projections under missing context]{\includegraphics[trim={0.2cm 0.5cm 0cm 0.5cm},clip,width=0.48\linewidth]{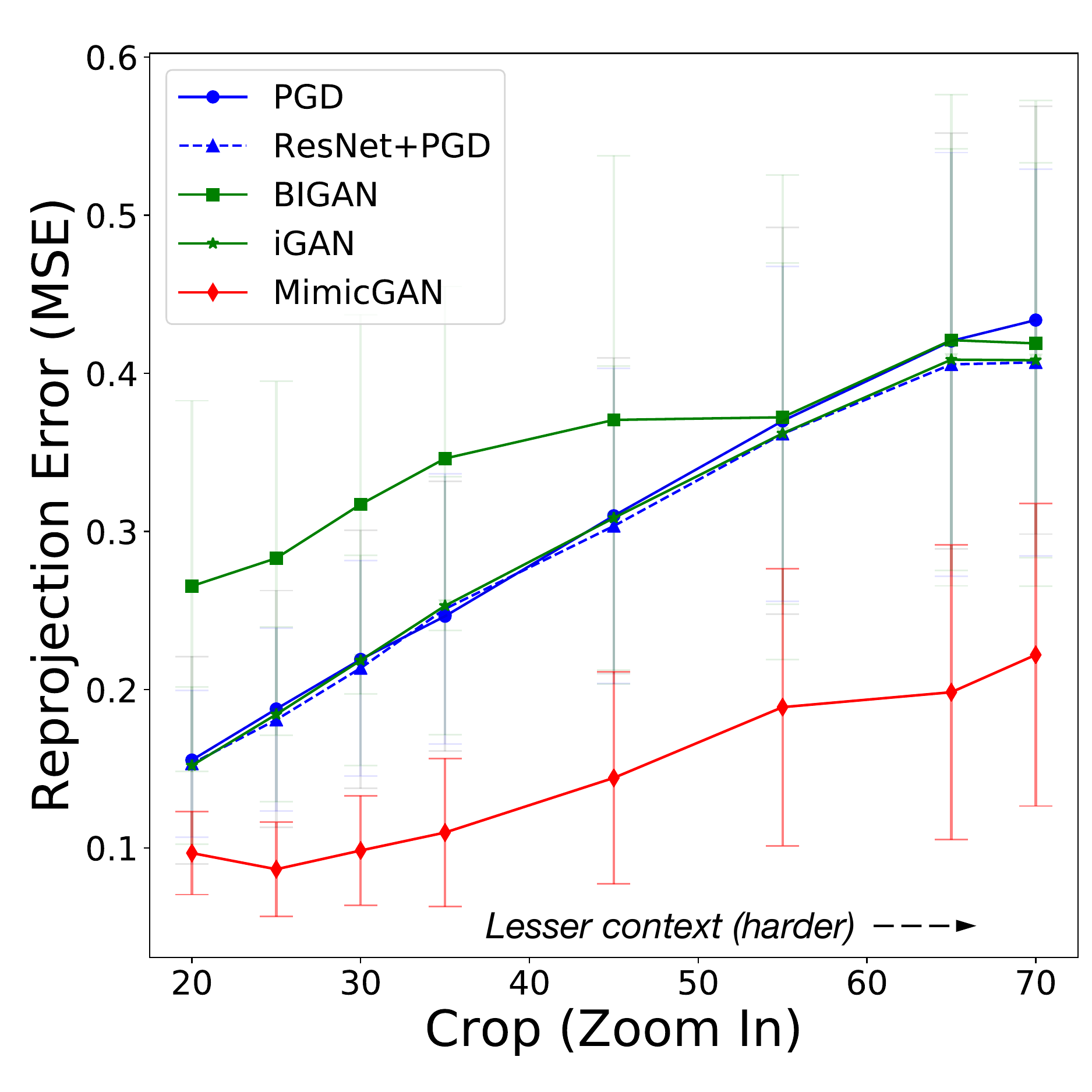}
\label{fig:zoom}}
\caption{\textbf{Reprojection error:} is shown here for 100 test images, along with its standard deviation. In all the four challenging cases considered here, \suttl~ significantly outperforms the baselines, which are widely used to explore the latent space of a GAN.}
\label{fig:reproj}
\end{figure*}

\begin{figure*}[!htb]
	\centering
\subfloat[Robustness to rotations (arrow indicates setting used during training)]{\includegraphics[trim={0.0cm 0.0cm 0.0cm 0.0cm},clip,width=0.7\linewidth]{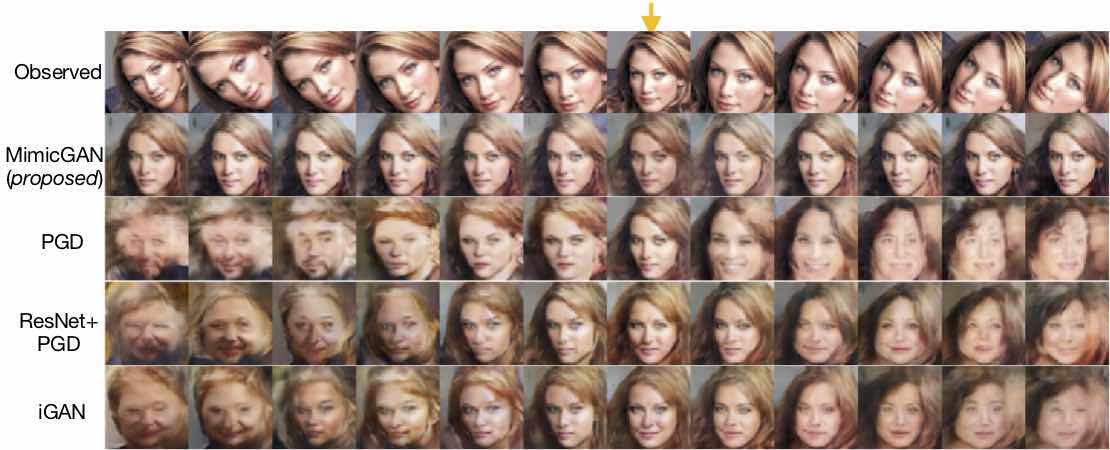}\label{fig:rot_inv}}

\subfloat[Robustness to scale (arrow indicates setting used during training)]{\includegraphics[trim={0.0cm 0.0cm 0.0cm 0.0cm},clip,width=0.7\linewidth]{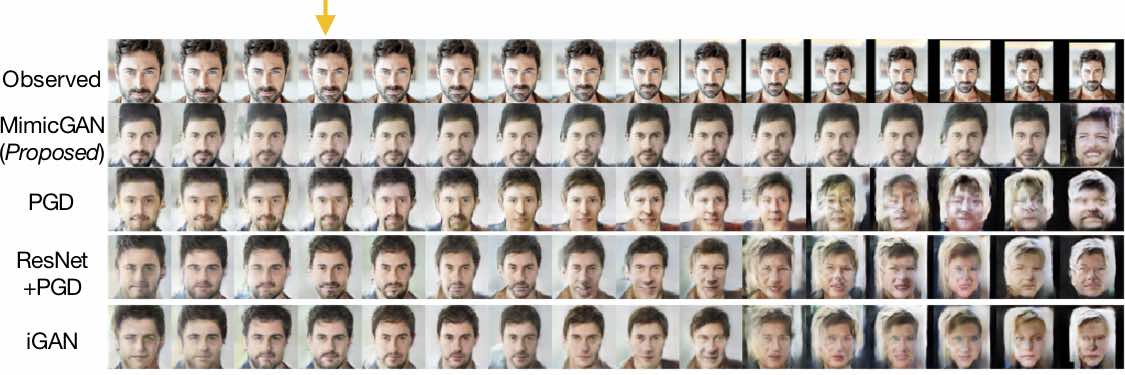}\label{fig:crop_inv}}

\subfloat[Robustness to missing pixels.]{\includegraphics[trim={0.0cm 0.0cm 0.0cm 0.0cm},clip,width=0.75\linewidth]{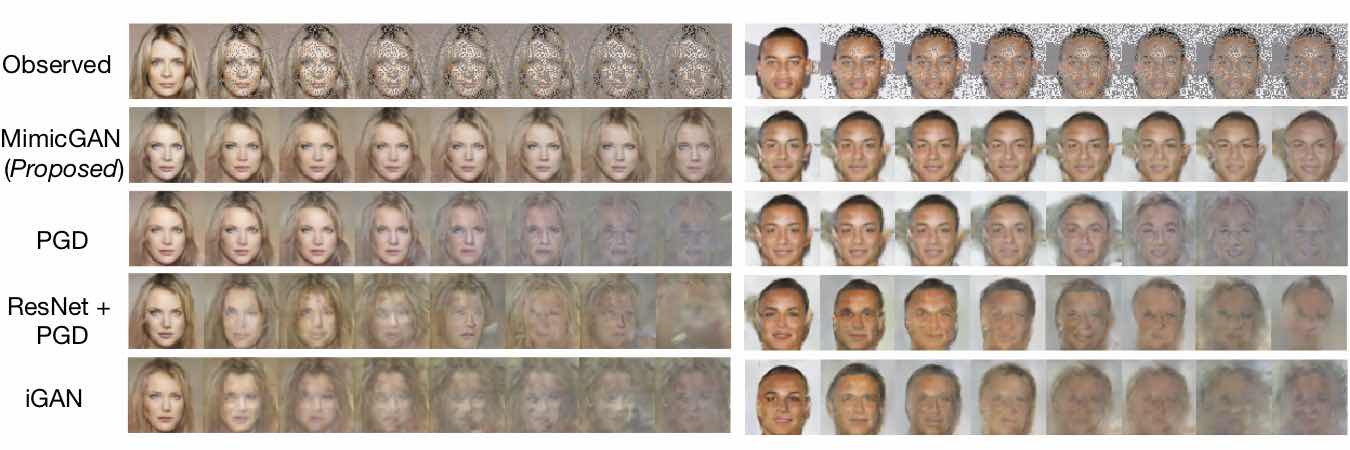}\label{fig:missing_fig}}

\subfloat[Robustness to missing contextual information.]{\includegraphics[trim={0.0cm 0.0cm 0.0cm 0.0cm},clip,width=0.75\linewidth]{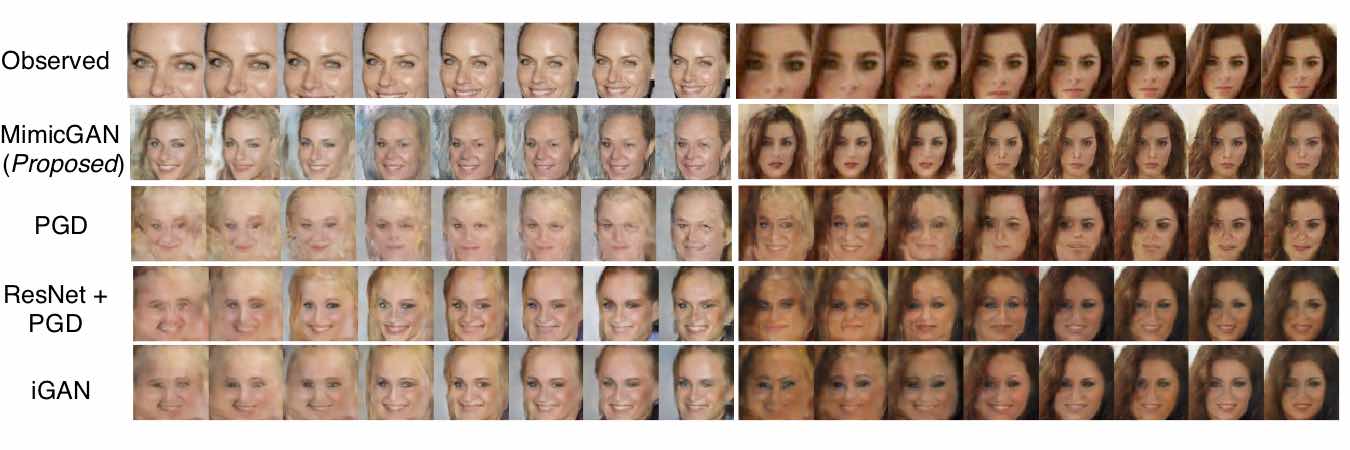}\label{fig:zoom_fig}}

\caption{Qualitative results demonstrating the invariance properties for different test images under all the corruptions considered here. Here, each image is from a different run with the specified corruption setting, and we see that \suttl~ provides significantly better reprojection error in all cases above. }
\label{fig:reproj_fig}
\end{figure*}

In this section, we demonstrate how corruption-mimicking with \suttl~improves projection quality across various deformations and transformations of a held-out test set, not accessed by the GAN during training. As described in the previous section, we assume that the GAN has been trained on \textit{clean} or un-corrupted data. In addition, we expect the corrupted images to be available only at test time, with no prior knowledge on the corruption function. For all empirical evaluation in this section, we measure the projection error as: 
\begin{equation}
\label{eq:proj_error}
\hspace{50pt} d_{proj} = ||X_{gt} - \mathcal{G}(\hat{z^*})||_2,
\end{equation}where $X_{gt}$ is the ground truth (i.e., test images with no corruptions, like exactly matching crop and rotation settings at train time) and $\hat{z^*}$ is the latent vector obtained upon convergence of the proposed algorithm \ref{mainAlg}. Next, we describe the different transformations and distortions considered for this study, along with experiment details. 

\paragraph{\textbf{Experiment Setup}:} We perform all our robustness experiments on the CelebA Faces dataset~\cite{liu2015faceattributes} which contains $202,599$ images. We crop each image by $147$ pixels, from the center of the image, and resize each image to be of size $64\times64$. We use the DCGAN~\cite{radford2015unsupervised} architecture, which is trained on $90\%$ of the images, while the rest $10\%$ are used for evaluation. Following~\cite{radford2015unsupervised}, we rescale all the images to be in the range of $[-1,1]$. Finally, we train the GANs on the original, clean images, and introduce corruptions only at test time.  Note, all our experiments were carried out using TensorFlow \cite{abadi2016tensorflow}. To corrupt the samples in each experiment, we draw random samples from the test set, followed up applying the same corruption on all of them, for simplicity. \suttl~ can work even when each sample is corrupted with slightly different parameters, where for e.g., each image is rotated by $\Theta_f \sim \mathcal{N}(\mathbf{\mu}_f,\mathbf{\sigma}_f)$. We test \suttl~ in this more challenging setting with adversarial defense experiment in section \ref{sec:adv}, and show it is still robust.

\paragraph{\textbf{Hyperparameters:}} In general, we find \suttl~to be generally stable to hyper-parameter choices in algorithm \ref{mainAlg}. In particular, the results reported were obtained with the following settings: $T_1 = 25, T_2 = 25, \gamma_s = 1e^-2, \gamma_g = 1e^-2$. Though these can be refined for different datasets, we observed these settings to produce consistent performance in all cases. It is important to note that in all the following experiments, the held-out set is directly presented to \suttl~ without any additional tuning or training.

\paragraph{\textbf{Baselines}:} Following the state-of-practice, we compare \suttl~against these baseline techniques: 
\begin{itemize}[leftmargin=*]
  \setlength\itemsep{1em}

\item[]\textbf{Projected Gradient Descent (PGD)}: This is a special case of \suttl~when the surrogate, $f \approx \mathbb{I}$ is identity. We fixed the learning rate for projected gradient descent at $5e-3$. Note that, PGD has been successfully used in several applications involving GANs~\cite{AsimDeblurGAN,bora2017compressed,shahICASSP2018,yeh2017semantic}. The optimal projection is given by $\mathcal{G}(z^*)$, where $z^* = \underset{\{z \in \mathbb{R}^{d}\}}{\arg\min} ||Y_{obs} - \mathcal{G}(z)||$, where we drop the sample index for convenience. This is the same as described in eq \eqref{eq:pgd}.

\item[]\textbf{ResNet + PGD}: We repeat PGD, but this time compute loss in the feature space from a pre-trained ResNet-50 \cite{he2016deep}. A version of this idea with VGG features has been used in \cite{hoshen2018non}, but we observed improved performance with ResNet features compared to VGG. We expect this to be a better baseline than simply using raw pixel intensities, since the ResNet has been trained on ImageNet \cite{russakovsky2015imagenet}, that include a wide range of image variations. However, it has been previously reported that even complex models such as the ResNet fail to generalize to even small perturbations \cite{azulay2018deep}. The optimal projection in this case is given by $\mathcal{G}(z^*)$, where $z^* = \underset{\{z \in \mathbb{R}^{d}\}}{\arg\min} ||\mathcal{R}(Y_{obs}) - \mathcal{R}(\mathcal{G}(z))||$, where $\mathcal{R}(x)$ extracts ResNet-50 features corresponding to the image $x$. We follow \cite{zhu2016generative}, and include a mixture of losses in both pixel space (like PGD) and ResNet feature space with a weight of $1e-3$.

\item[]\textbf{iGAN} \cite{zhu2016generative} : Interactive-GAN learns an encoder on the fly to determine the best projection. We use an encoder with similar architecture as the discriminator except the last layer that contains the appropriate number of neurons as the latent dimension. This network, $E(Y^{obs},\theta_E): \mathcal{I}\mapsto \mathbb{R}^d$ is trained given observations $Y^{obs}$ such that the optimal projection is given by: $\mathcal{G}(E(Y^{obs},\theta_E^*)), $ where $\theta_E^* = \underset{\theta_E}{\arg\min} ||\mathcal{G}(E(Y^{obs},\theta_E)) - Y^{obs}||$.

\item[]\textbf{BIGAN/ALI} \cite{donahue2016adversarial}, \cite{dumoulin2016ALI}: This is a technique that learns an encoder to map into the latent space directly, while training the GAN. Considering our final goal is to project onto the image manifold, one can use the learned encoder directly. The optimal projection is obtained by passing the image through the encoder and the decoder directly as $\mathcal{G}(\mathcal{E}(Y^{obs})),$ where $\mathcal{E}$ is the encoder in the BIGAN.
\vspace{10pt}

In all cases, we measure the quality of projection as the error between the original image, and the one obtained from the generator after encoding. We also  use the same pre-trained GAN in all the baselines, with the only factor of variation being the projection technique.
\end{itemize}

\subsection{\textbf{Robustness to Affine Transforms}} In this experiment, we study how \suttl~  can be used to obtain scale and shift invariant projections on the image manifold. These form the most common type of corruptions one can expect to observe in the wild. We consider two such transformations: (a) scale: where we provide more context of the image (``zooming out'') than what the GAN has observed during training, and (b) rotation: here we rotate the images with a small crop to avoid edge artifacts. As we will demonstrate, even these seemingly simple transformations can significantly throw-off modern deep learning systems, unless they have appeared in the training set \cite{azulay2018deep}.

\suttl~provides robustness to these transformations over a wide range as shown in Figures \ref{fig:crop} and\ref{fig:rotation} respectively. We chose $100$ random examples from the held-out test set, and compared the projection error (mean and standard deviation) obtained for different amounts of scaling or rotation. It is obvious from these plots that the baseline techniques are extremely sensitive to even small changes in these settings. In comparison, \suttl~ demonstrates invariant properties over a large magnitude of perturbations. Additionally, in figures \ref{fig:crop_inv} and \ref{fig:rot_inv} we show examples of all crops and rotations on a particular test image, and demonstrate the effectiveness of~\suttl. Note the canonical setting used during the training is marked with an arrow. An interesting observation from these figures is that all baseline methods fail somewhat differently.
\subsection{\textbf{Projections with missing or partial information}}
Next, we study the robustness of \suttl~ to missing information or lack of complete context while projecting onto the manifold. Specifically, we consider two commonly observed cases: (a) Missing pixels: Here we randomly drop pixels from the image, with the number of missing pixels varied between $0-60\%$. (b) Zoom-in: We intentionally leave out a lot of context from the original image by cropping very close to the face. In both these cases, obtaining an accurate projection is an ill-posed inverse problem since there can be infinitely many solutions for the same observation, especially in cases when the corruption is extreme. We show that, even in this ill-posed setting, \suttl~ is able to provide meaningful and highly plausible projections. 

Projection results obtained under these corruptions are showed in Figures \ref{fig:missing_fig} and \ref{fig:zoom_fig} respectively. Quantitative results obtained by averaging the errors over $100$ randomly chosen test examples are shown in Figures \ref{fig:missing}, and \ref{fig:zoom}.  We observe that while BIGAN fails easily with even small amounts of missing pixels. On the other hand, PGD appears to be more stable, but eventually fails beyond $40\%$ missing pixels. In comparison, \suttl~ remains robust at a much higher level of corruption, and produces lower projection errors.

\begin{figure*}[!htb]
\centering
\subfloat[\color{blue}Quality of projection under different number of observations. The high variance in projection with just a single observation is due to both an overfit surrogate and the \emph{identifiability issues} arising from not being able to tell apart the underlying true signal from the corruption]{
\includegraphics[trim={0.0cm 1.5cm 0.0cm 0.0cm},clip,width=0.6\linewidth]{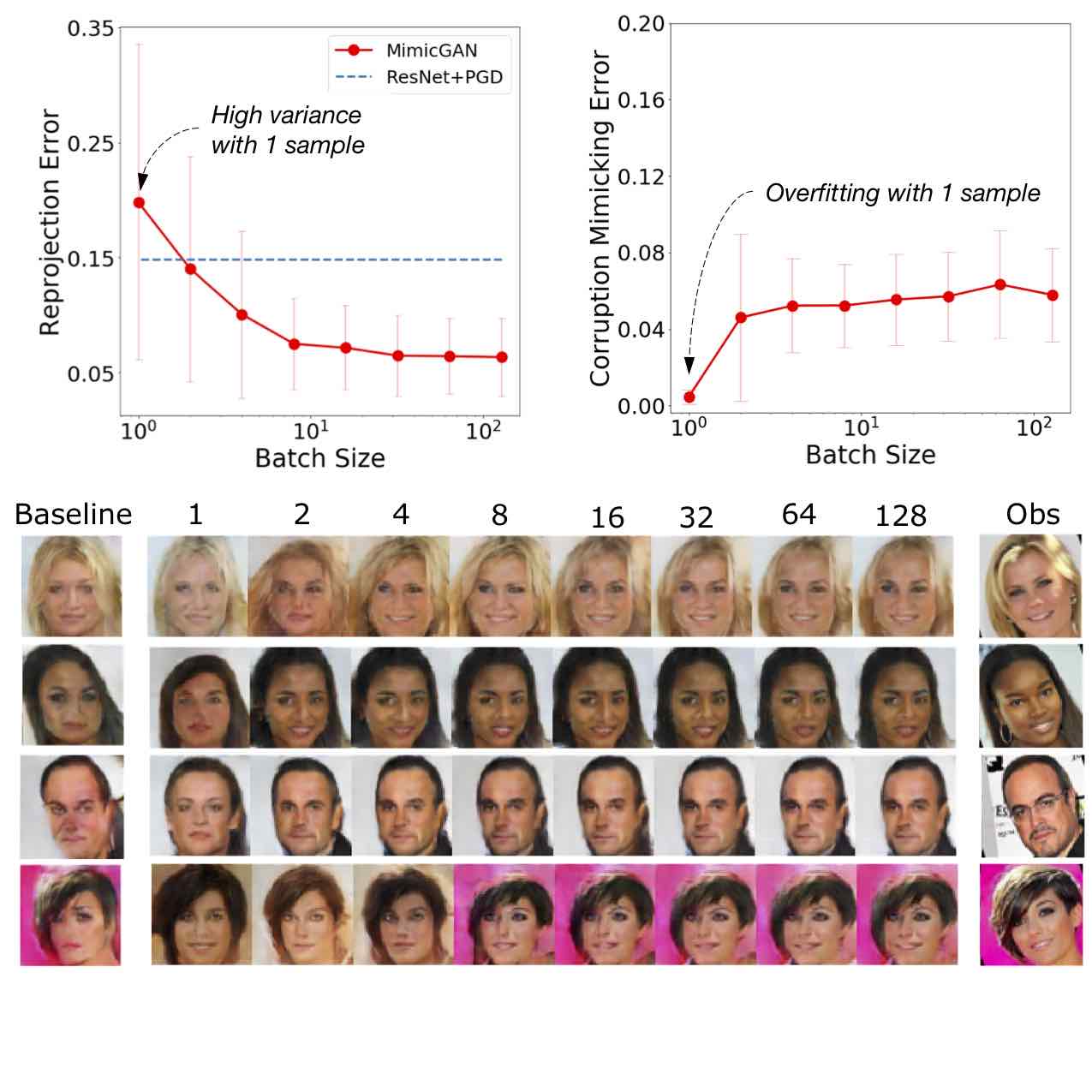}
\vspace{-30pt}
\label{fig:properties}}
\vspace{-10pt}

\subfloat[Robustness on FFHQ dataset with $128\times 128$ images]{\includegraphics[trim={0.2cm 0.5cm 0cm 0.5cm},clip,width=0.4\linewidth]{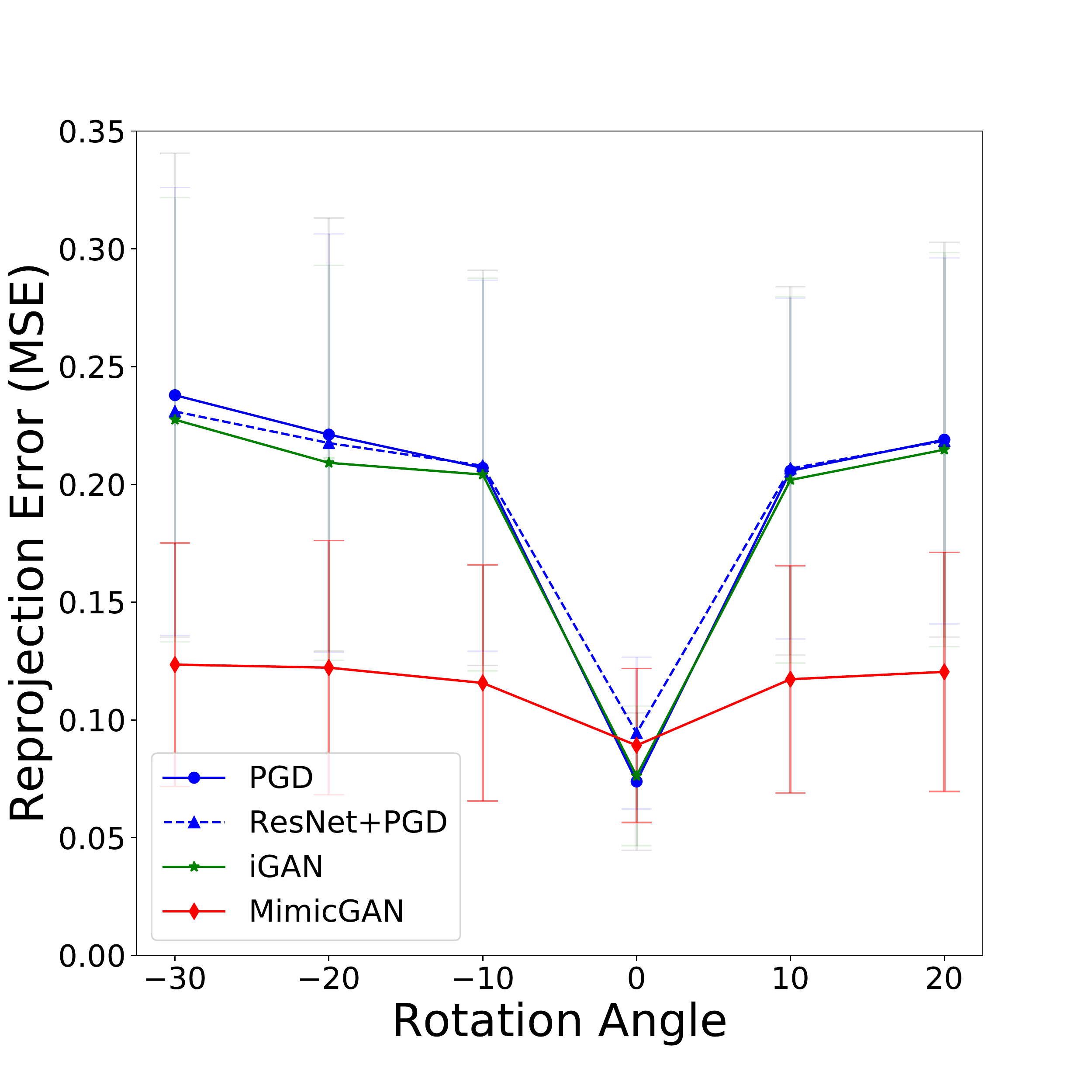}
\label{fig:128}}
\subfloat[Time complexity per 100 images]{\includegraphics[trim={0.0cm 0.0cm 3.5cm 2.5cm},clip,width=0.4\linewidth]{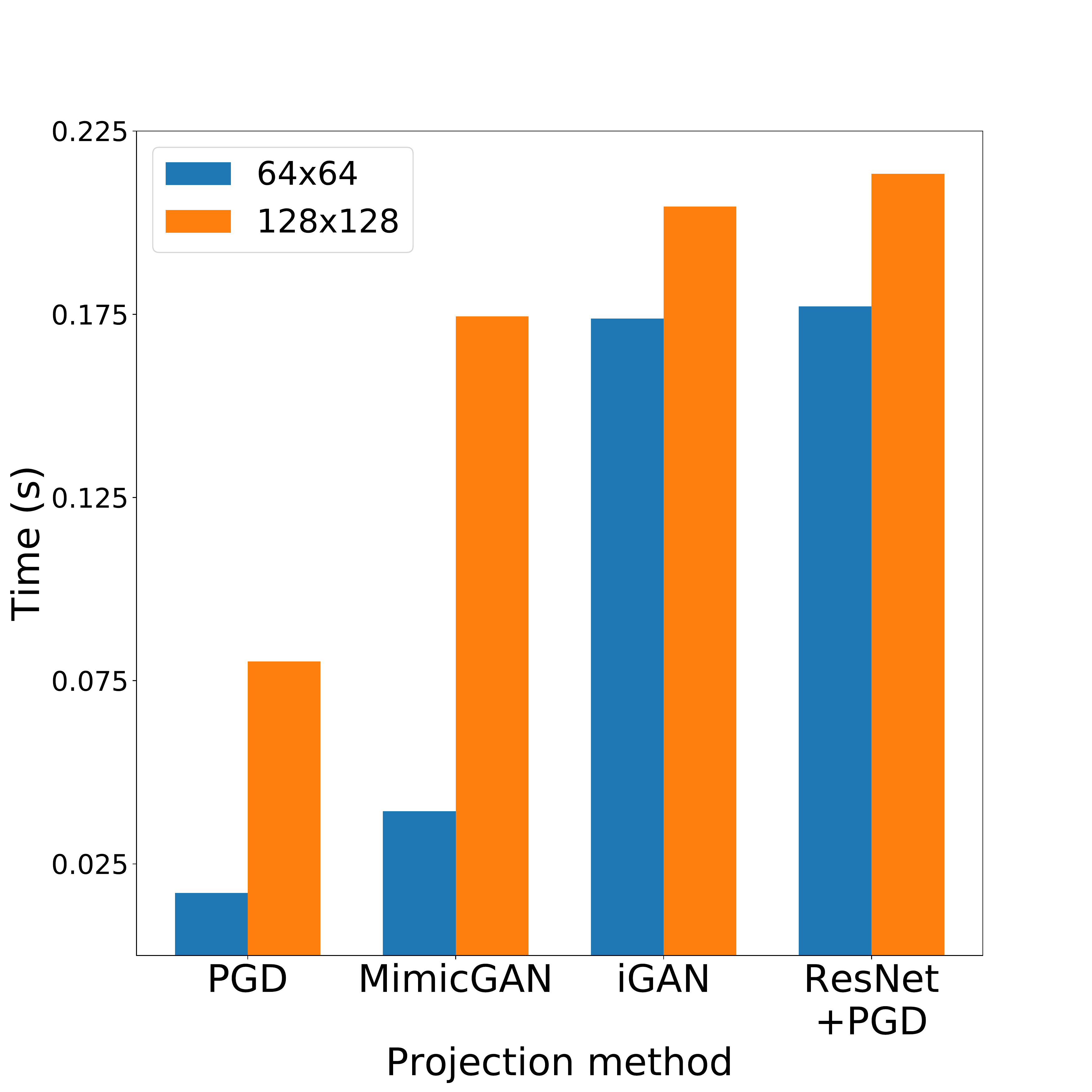}\label{fig:time}}
\caption{\color{blue} \textbf{Properties of \suttl:} (a) Number of observations vs quality of projection, and corruption mimicking. (b) Robustness properties generalize to larger images and (c) Time taken to compute the projection compared to different baselines.}
\label{fig:prop_all}
\end{figure*}

\subsection{\textbf{Properties of \suttl}}
\label{sec:properties}
Finally, we study how \suttl~ performs under different conditions when the number of observations vary, computational time, and how it generlizes to larger images of size $128\times128$.
\paragraph{\textbf{Relationship to number of observations}} Since the surrogate network is central to \suttl, the number of available observations is expected to play an important role in the quality of projection. To quantify this relationship, we take a total of 128 images from the CelebA dataset and project them using \suttl~by varying the batch size as follows -- [1, 2, 4, 8, 16, 32, 64, 128]. We compare two metrics: quality of reprojection (same as eqn. \eqref{eq:proj_error}) and corruption mimicking error, which is the same as the optimization cost defined in equation \eqref{eq:obj}. Ideally, we want both to be as small as possible. 

\noindent In Figure \ref{fig:properties}, we show the results for this experiment, where the corruption considered is rotation. We observe that, as expected with more observations the quality improves. However, we note that 8-10 samples are typically sufficient to recover the same projection error as 100 observations. Moreover, with a single observation the surrogate overfits leading to low corruption mimicking error, but a high reprojection error. The high variance in the reprojection error is explained by the fact that when only a single observation is available, we run into an identifiability issue -- i.e., \suttl~ can no longer distinguish what the corruption was from the original image. We note we obtain better reprojection error, and higher visual quality of images even with just 2 observations when compared to a competitive baseline, which is ResNet+PGD.

\paragraph{\textbf{Time Complexity Trade-Off}} Naturally, using a surrogate network leads to a computational burden when compared with vanilla PGD because of training the surrogate. In Figure \ref{fig:time}, we show the time taken to project 100 random samples for images of size $64\times64$, and $128\times128$. We compute the time taken per iteration for each of the methods, on the same NVIDIA P100 GPU with 16GB of memory. We observe that, as expected, \suttl~ tends to take $\sim2$x the time than PGD, but is significantly faster than iGAN or ResNet+PGD in both sizes of images. This is primarily because computing ResNet features becomes the bottleneck in these methods.

\paragraph{\textbf{Generalization to larger images}} We also show that robust projection can be achieved with larger images, using exactly the same surrogate model (adjusted for size) and hyper parameter settings. We see similar robustness advantages even for these images. In Figure \ref{fig:128}, we show how \suttl~ provides robust projections at $128\times128$ sized images, when the corruption is an unknown rotation.
\section{Applications of Robust Projection}
\begin{figure*}
	\centering
	\subfloat[\textbf{CELEBA$\rightarrow$LFW}: MimicGAN robustness to distribution shifts as seen here, where the standard projected gradient descent (PGD) completely fails to recover accurate projections.]{\includegraphics[trim={0.0cm 0cm 0.0cm 0.0cm},clip,width=0.75\linewidth]{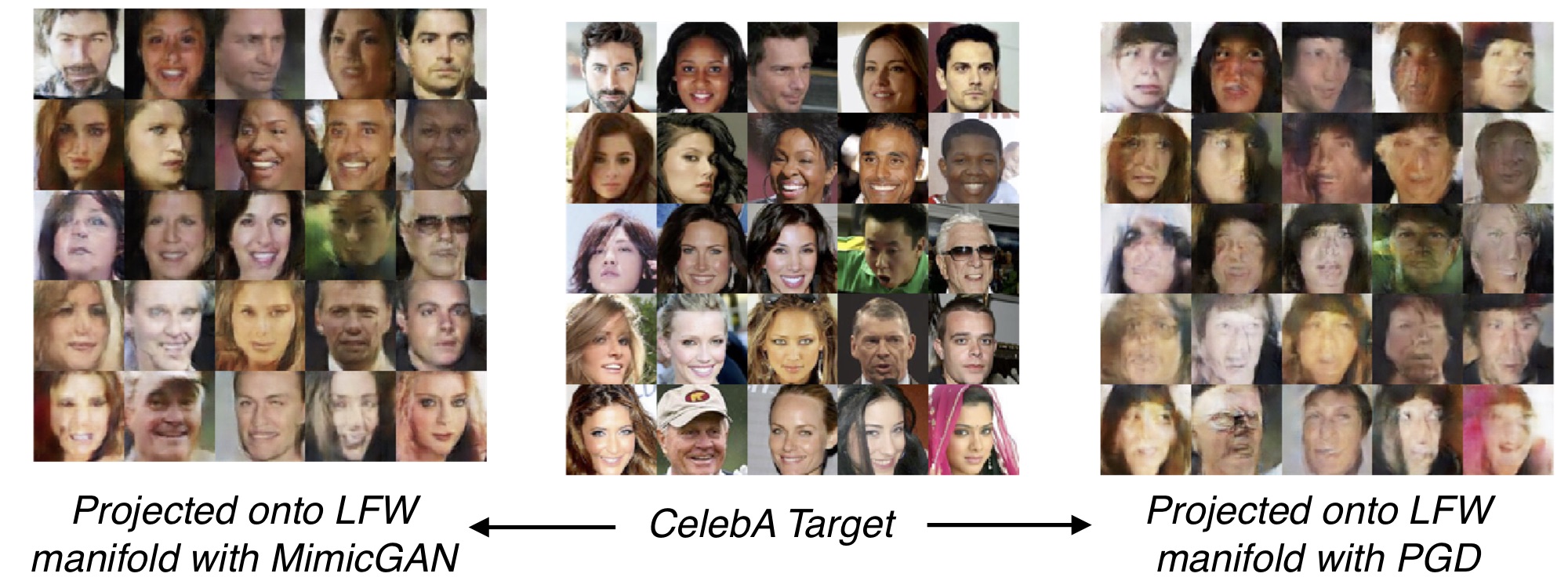}\label{fig:celeba2lfw}}
	
	\subfloat[\textbf{Mix 'n Match:} MimicGAN provides the flexibility to use different GANs for datasets other than what was used during training.]{\includegraphics[trim={0.0cm 2.0cm 2.0cm 0.0cm},clip,width=0.75\linewidth]{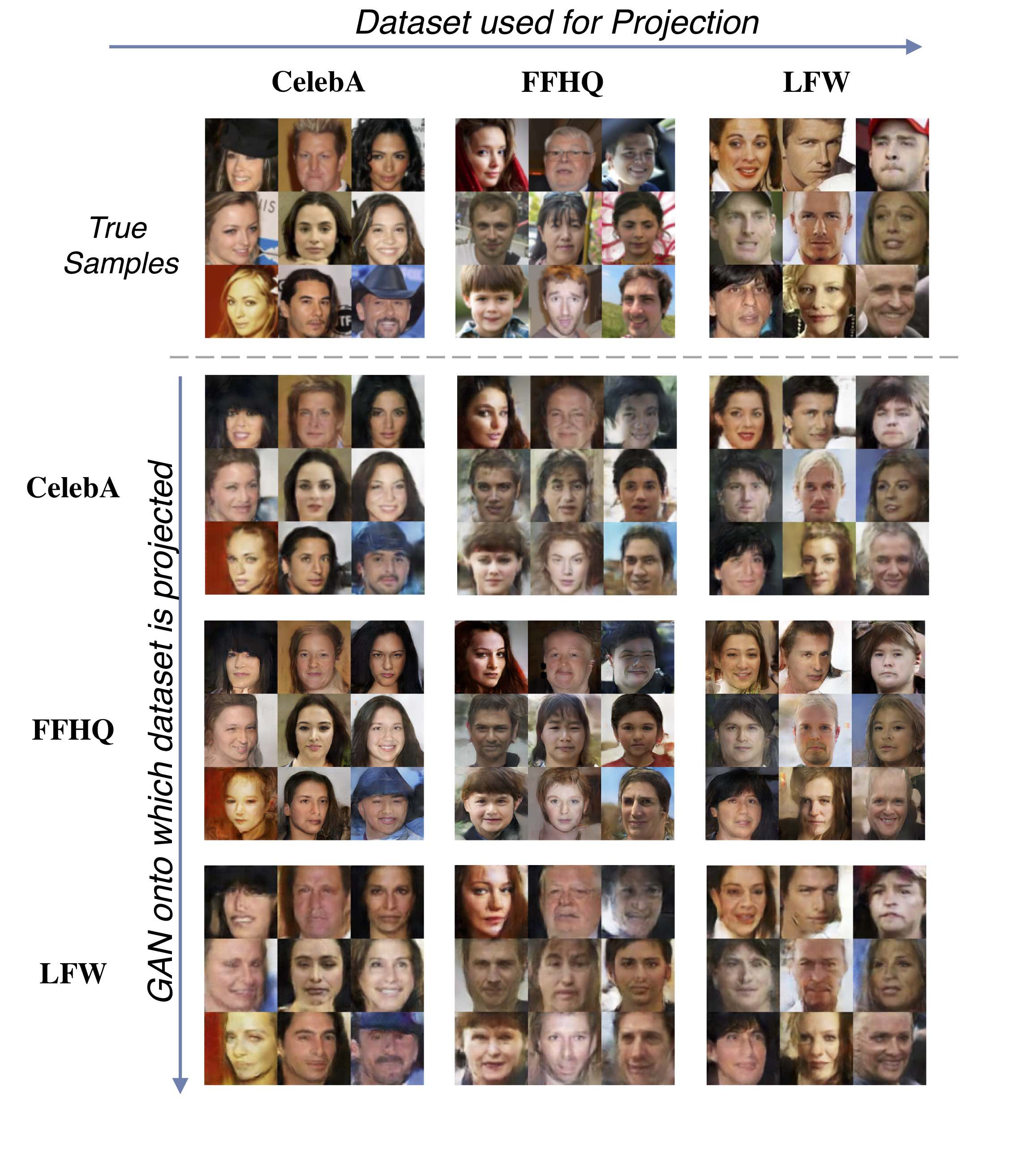}\label{fig:mixnmatch}}
	\caption{\color{blue} Here we show projections obtained with \suttl, i.e. $\mathcal{G}(z^*)$, from three datasets CelebA \cite{liu2015faceattributes}, FFHQ \cite{karras2018style}, and LFW \cite{pinto2011scaling}, where held out samples are projected onto GANs trained on each of these datasets. The images to be projected are randomly chosen from each of the datasets. It can be seen that \suttl~ provides very meaningful projections across distributions, without explicitly knowing how these datasets are related to each other.}
\end{figure*}
In this section, we study how robust projections can lead to performance gains in different applications involving GANs. The idea of projecting onto a known image manifold has been leveraged in applications such as adversarial defense \cite{defenseGAN,ilyas2017robust,santhanam2018defending}, anomaly detection \cite{zenati2018efficient,akcay2018ganomaly,an2015variational}, domain adaptation, etc. We use the following experiments to verify our hypothesis that a more robust projection can make GAN based solutions more effective in these applications. 

\subsection{\textbf{Distribution Alignment}}
An important application of GANs has been in the context of handling distribution shifts across datasets, and we evaluate the effectiveness of ~\suttl~ in solving this challenging problem. First, we consider the problem of projecting face images onto a manifold inferred using face datasets characterized by systematic distributional shifts. For example, projecting a child's face using GANs trained extensively using adult faces is quite challenging in practice. Subsequently, we consider the problem of unsupervised domain adaptation, which attempts to leverage labeled data from a source dataset to build a classifier for an unlabeled target dataset, when there is an unknown distributional shift between the two datasets.

\subsubsection{Adapting Face GANs}
In this application, we consider three different face image datasets, namely CelebA,  LFW~\cite{becker2013evaluating} and FFHQ~\cite{karras2018style}, and evaluate the quality of projecting faces from one dataset onto GANs trained using another dataset. With existing techniques, the projection operation can completely fail unless we know \emph{a priori} how to normalize the distribution shift, which is challenging in practice. Instead, we show how \suttl~ can help in accounting for this distribution shift such that any relevant GAN ``backend'' can be used without loss of functionality. 

\paragraph{ Datasets:}
\begin{itemize}
	\item[1.] \textbf{LFW:} A combination of PubFig83 \cite{pinto2011scaling} and Labelled Faces in the Wild (LFW) aligned dataset \cite{LFWTech}, following \cite{becker2013evaluating}. This dataset contains 25,068 images of celebrities. We use the original settings proposed in this paper, and use a crop of 200 that only contains the close up of faces, followed by resizing them to be $64\times 64$.
	\item[2.] \textbf{FFHQ:} \cite{karras2018style} contains 70,001 images scraped from Flickr, at multiple resolutions, and we use the thumbnail version with images of size $128\times 128$ without any additional crops or resizing.
\end{itemize}

Note that, for all datasets, we used a DCGAN architecture similar to the one described in Section \ref{sec:da}. The only difference was in the case of FFHQ, wherein we included an additional layer to both the generator and discriminator to account for the change in image size ($128\times 128$). In Figure \ref{fig:mixnmatch}, we illustrate the results obtained by projecting CelebA faces onto an LFW GAN. For comparison, we also show the corresponding projections obtained using the baseline PGD in figure \ref{fig:celeba2lfw}. It is seen that PGD tries to identify images that exactly match the given inputs, which, if the mode doesn't exist in the distribution, result in sub-optimal projections. In contrast, \suttl~ produces significantly higher quality projections by automatically accounting for the distribution shifts. As applicable, we either upsample or downsmple the source images to maintain the same size as the target dataset.

\subsubsection{{Unsupervised Domain Adaptation}}

\label{sec:da}
The task considered here is the alignment between two related yet distinct domains (source and target), while building a classifier for the target without access to any labeled data. For this experiment, we use the commonly benchmarked pair of domains in handwritten digit recognition, namely MNIST and USPS. We argue distribution alignment can be seen as essentially projecting the target data onto a source data manifold. This task is challenging due to unknown variations in scale, skew, rotation, and other statistical properties across these two domains, causing existing projection techniques to fail. In contrast, MimicGAN can project varied datasets onto each other with ease, which we exploit for the problem of unsupervised domain adaptation. We follow the experiment setup in several domain adaptation works in the literature, and show its effectiveness under a 1-NN and a more complex CNN classifier. 

\begin{figure*}[t]
	\centering
	\subfloat[USPS$\rightarrow$ MNIST]{\includegraphics[trim={0.0cm 0cm 0.0cm 0.0cm},clip,width=0.85\linewidth]{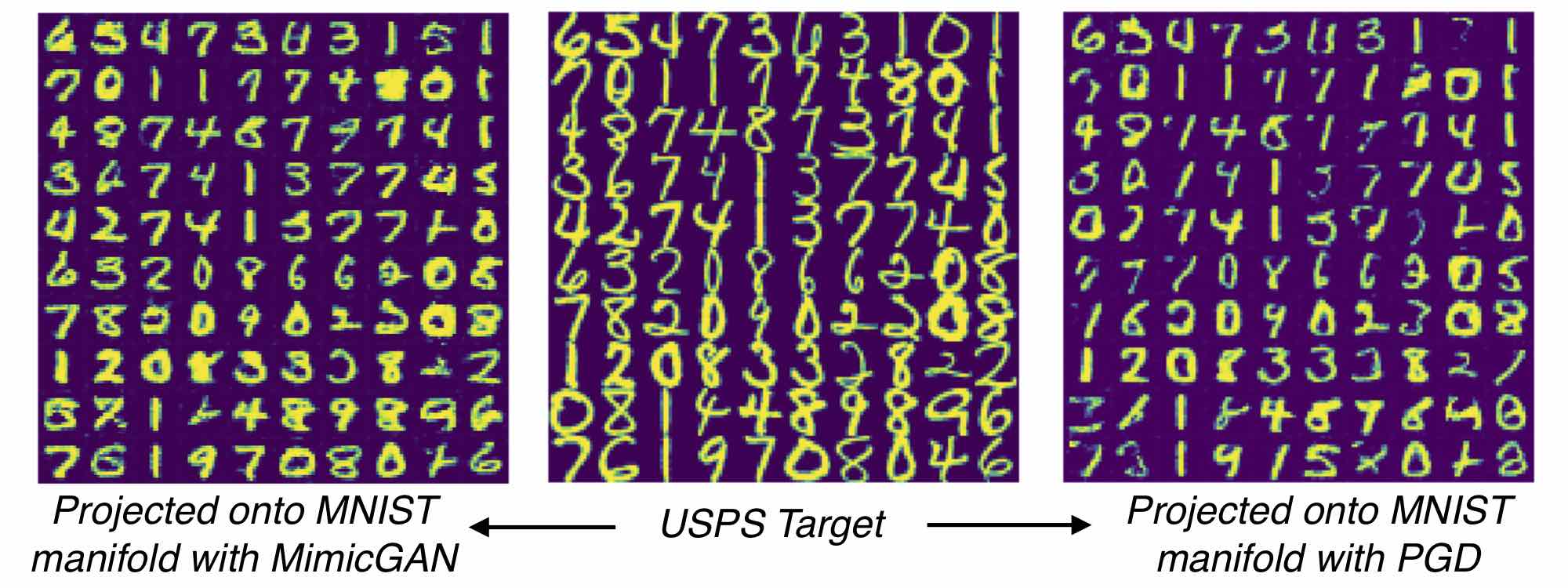}\label{fig:da_usps2mnist}}\quad
	\subfloat[MNIST$\rightarrow$USPS]{\includegraphics[trim={0.0cm 0cm 0.0cm 0.0cm},clip,width=0.85\linewidth]{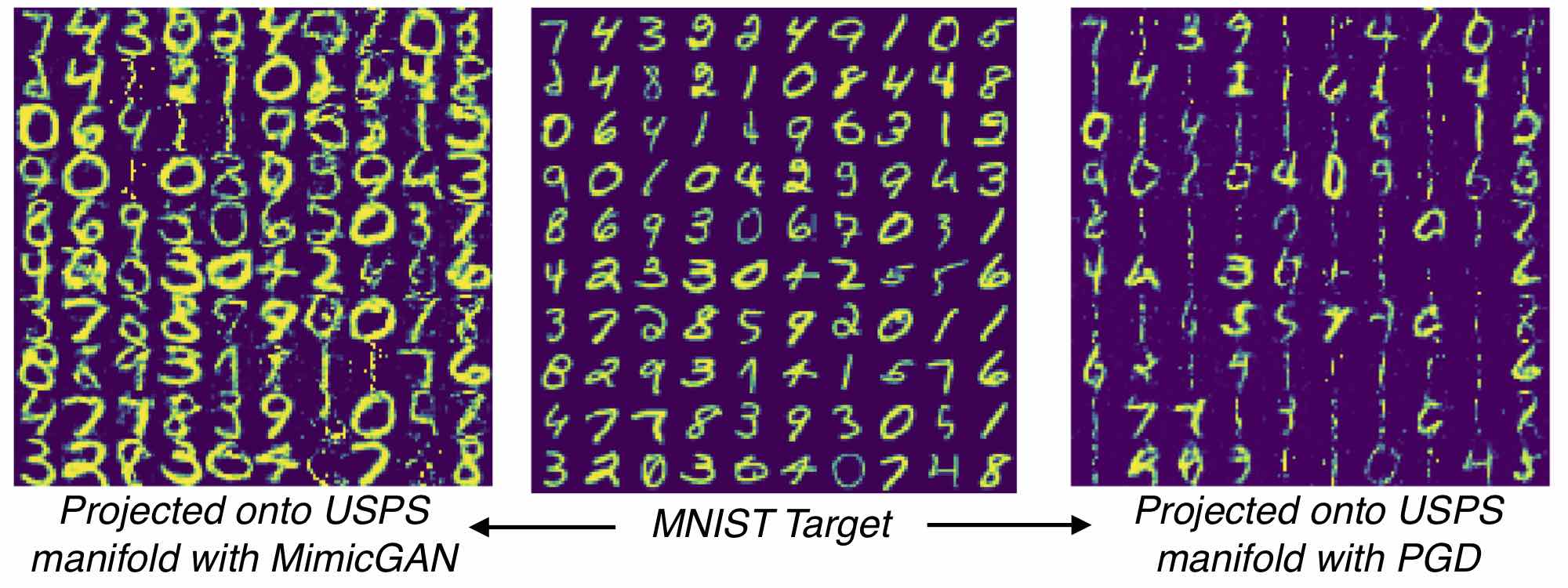}\label{fig:da_mnist2mnist}}\quad
	\caption{\color{blue} MimicGAN allows for accurate projection under distribution shifts, as shown here with MNIST and USPS datasets. The images to be projected are randomly chosen from the other dataset, and we show $\mathcal{G}(z^*)$ here. The domain adaptation results on the full 65000 MNIST and 9298 USPS digits are shown in Table \ref{tab:daGAN}.}
\end{figure*}

\noindent \textbf{Setup:} We follow the setup in~\cite{seguy2017large}, where we perform adaptation on the entire 65000 digits of MNIST and 9298 digits of USPS. We resized the MNIST digits to 16x16 so they are the same size as the latter, and used the following hyper-parameter settings: $U\rightarrow M: T_1 = 25, T_2 = 75,\gamma_s = 1e-2, \gamma_g = 8e-2$, and $M\rightarrow U: T_1 = T_2 = 15, \gamma_s=1e-2, \gamma_g=9e-3$. We perform the projection in batches of size $1000$ at a time for efficiency. These hyper-parameters are not very sensitive, and are chosen here based on a validation set of 100 images.
\vspace{5pt}

\noindent \textbf{Baselines:} We compare the performance of \suttl~ in distribution alignment with the following baselines:
\begin{itemize}
\item No Adaptation: Here we use a nearest neighbor classifier trained on the source dataset directly on the target dataset. 
\item PGD: We compare performance when no corruption-mimicking is performed, with the same GAN backend used by \suttl~.
\item Subspace Alignment \cite{fernando2013unsupervised} A classical domain adpatation approach that aligns two distributions based on a single subspace that minimizes the Forbenius norm between the target and the subspace aligned source. The source aligned coordinate system is given by $X_a = X_SX_S'X_T$, where $X_S$ is the subspace fit to the source dataset, and $X_T$ the subspace of the target dataset.
\item Large Scale Optimal Transport (OT) \cite{seguy2017large} A recent technique that leverages an efficient version of optimal transport to align large datasets.
\end{itemize}
{\color{blue}
Additionally, we compare the performance of \suttl~to more recent domain adaptation approaches in Table \ref{tab:daGAN}. For this we follow standard protocol and train a CNN on the source dataset and test it on the aligned target dataset, we verify that our \emph{no adaptation} numbers are very close to those reported in most recent papers.} 
\vspace{5pt}

\begin{table*}[t]
	\color{blue}

	\centering	
	\begin{tabular}{|c|lp{0.8in}p{0.8in}|c|}
	
		\hline
		\multicolumn{1}{|c}{} & \multicolumn{1}{l}{\textbf{Method}} & \multicolumn{1}{p{0.8in}}{\textbf{M$\rightarrow$ U}}& \multicolumn{1}{p{0.8in}}{\textbf{U$\rightarrow$ M}}&\multicolumn{1}{c|}{}\\\hline
		 \parbox[t]{1mm}{\multirow{7}{*}{\rotatebox[origin=c]{90}{\emph{1-NN}}}} &&&&\parbox[t]{1mm}{\multirow{8}{*}{\rotatebox[origin=c]{90}{{\color{red}Task Agnostic}}}}\\
		& No Adaptation & 71.26 & 31.65& \\
		& PGD  & 58.15& 24.96& \\
		& \small{Subspace Alignment} \cite{fernando2013unsupervised} &59.53 & 44.61&  \\
		& Large Scale OT \cite{seguy2017large} & 77.92 & 60.50& \\
		& MimicGAN (\emph{ours}) &{80.75} &\textbf{\color{red}65.50}&\\ \cdashline{1-4}
		\parbox[t]{2mm}{\multirow{9}{*}{\rotatebox[origin=c]{90}{\emph{CNN }}}} & & & &\\
		& No Adaptation & 81.6 & 60.3&  \\
		& PGD & 61.7 & 48.2&  \\
		& MimicGAN \emph{(ours)} & \textbf{\color{red}83.7} & 63.3& \\\cdashline{2-5}
	& & & & \parbox[t]{2mm}{\multirow{6}{*}{\rotatebox[origin=c]{90}{{Task Specific}}}}\\
		& CoGAN \cite{liu2016coupled} & 91.2 &  89.1 & \\
		& ADDA \cite{tzeng2017adversarial} & 92.4 & 93.8 & \\
		& Gen to Adapt \cite{sankaranarayanan2018generate} & 95.3 & 90.8 & \\
		& DANN \cite{ganin2016domain} & \textbf{95.7} & 90.0 & \\
		& CyCADA \cite{hoffman2017cycada} & 95.6 & \textbf{96.5} & \\
		
		\hline
	\end{tabular}
		\caption{An application of distribution alignment in unsupervised domain adaptation (UDA). MimicGAN outperforms many task-agnostic alignment strategies, i.e. methods that do not have access to source as well as target labels, using a pixel-based 1-nearest neighbor (1-NN) classifier, as well as a convolutional neural network (CNN). For context, we also compare with recent task-specific adaptation techniques.}

	\label{tab:daGAN}
\end{table*}

\noindent \textbf{Results:} We show the results of projection for both datasets in Figures \ref{fig:da_usps2mnist}, and \ref{fig:da_mnist2mnist}, and it is clearly evident that corruption-mimicking significantly improves the quality of projection as compared to PGD. We note that PGD never recovers the true samples from the target distribution, and only recovers those images very similar to those the GAN has previously seen. We also quantitatively evaluate the 1-NN classifier performance after performing the alignment using \suttl. Table \ref{tab:daGAN} shows that alignment under MimicGAN is superior to even large scale optimal transport \cite{seguy2017large}, which computes a sample-wise alignment between two entire distributions. More importantly, using \suttl~improves perfromance over PGD by nearly 20 and 30 percentage points in MNIST to USPS, and USPS to MNIST respectively. This significant boost is attributed solely to the improved quality of projection while accessing the target distribution directly at test time, making it a viable domain generalization technique. 

For the CNN based methods, it should be noted that the current state of the art methods in domain adaptation are all task-specific adaptation, i.e. the alignment is closely tied to the classifier, and uses labels from the source domain. On the other hand, \suttl~ aligns the distribution in a task-agnostic manner, i.e. with no knowledge of either the CNN classifier or the source labels. As a result, \suttl~ is inferior to the existing task-specific methods, as expected. However, it is worth noting that among the task-agnostic methods (pure distribution alignment), \suttl~ obtains the highest adaptation accuracy.


\begin{figure*}[!htb]

	\centering
	\subfloat[Universal Perturbations \cite{moosavi2017universal}]{\includegraphics[trim={0.0cm 0cm 0cm 0cm},clip,width=0.3\linewidth]{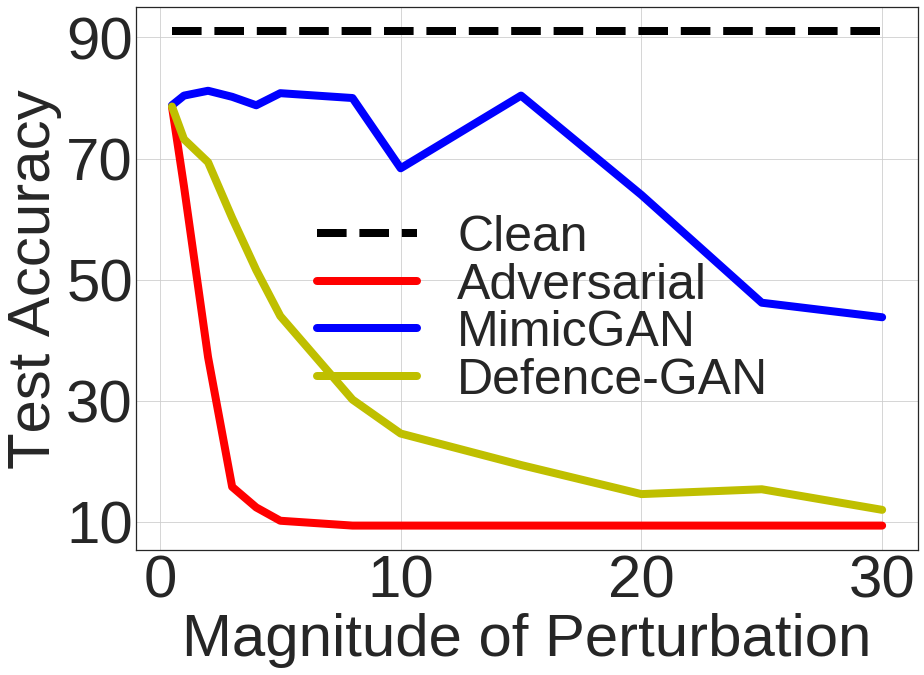}
		\label{fig:universal}}
		
	\subfloat[][\small{Popular Adversarial Perturbations}]{
	\color{blue}
		\centering
		\small
		\begin{tabular}{p{3.1cm}cccc}
			\hline
			
			\textbf{Attack} &No Defense& Cowboy \cite{santhanam2018defending}&Defense GAN  \cite{defenseGAN}& \suttl\hspace{1pt}\emph{(Ours)}\\ \hline
			BIM \cite{kurakin2016adversarial}			& 05.60   &  $66.00 \pm 3.47$ 	& $68.46 \pm 3.12$		& $\mathbf{72.20 \pm 3.80}$ \\
			DF \cite{moosavi2016deepfool}			& 06.20   &  $71.43 \pm 4.72$ 	& $78.86 \pm 1.36$ 	& $\mathbf{82.40 \pm 3.81}$ \\
			FGSM \cite{GoodfellowSS14}					& 11.60   &  $61.60 \pm 4.99$ 	& $65.93 \pm 2.44$		& $\mathbf{67.60 \pm 2.83}$		\\
						CWL \cite{CWL2}					& 00.40   &  $68.80 \pm 3.84$ 	& $71.67 \pm 3.16$ 	& $\mathbf{75.00 \pm 4.38}$\\
			PGDM \cite{madry2018towards}				& 05.40   &  $65.20 \pm 5.14$ 	& $69.90 \pm 2.17$ 	& $\mathbf{75.73 \pm 4.35}$ \\
			Obfuscated \cite{athalye2018obfuscated} (attack includes GAN)	& $26.93 \pm 4.29$ 	&  $27.73 \pm 4.50$ 	& $27.26 \pm 2.49$		& $\mathbf{31.06 \pm 4.01}$ \\ \hline
			
		\end{tabular}
		\label{fig:adv_table}
	}
	\caption{\textbf{Adversarial Defense:} \suttl~~defense shows significantly higher robustness to several strong adversarial attacks. Performance on clean test data is $91.5\%$. Here, we show the average defense on adversarial attacks obtained using 3 different GANs on the same dataset. Across the board we see that ~\suttl provides a stronger defense, even in the case of Obfuscated Gradients Attack \cite{athalye2018obfuscated}, where the GAN is included in designing the adversarial attack.}
	\label{fig:adv}
\end{figure*}

\subsection{\bf Adversarial Defense}
\label{sec:adv}
Here we study how~\suttl, by design, can provide effective defense against several state-of-the-art adversarial attacks. We argue that a robust projection onto the image manifold results in very effective cleaning of adversarial data. In this context, the \suttl~defense can be viewed as a generalization of the recent GAN-based defenses \cite{defenseGAN,ilyas2017robust,santhanam2018defending} that assume the corruption function $\hat{f}$ to be identity, similar to the PGD baseline. The more recent \emph{Cowboy} defense \cite{santhanam2018defending} uses an additional adversarial loss term to the defense-GAN loss~\cite{defenseGAN} with the discriminator (We implement it with $\lambda_{adv} = 1\mathrm{e}{-3}$). 

We consider a variety of strong attacks to benchmark our defense, and we find that in every single case the \suttl~defense is significantly stronger than existing techniques. While we outperform \textit{Defense-GAN}\cite{defenseGAN}, we retain its advantages -- i.e, the \suttl~defense is a test-time only algorithm that does not require any additional training. It is also entirely unsupervised, and does not need knowledge of the classifier prior to deploying the defense, thus leading to a practical defense strategy.

\paragraph{\textbf{Setup:}} We use a CNN classifier that achieves a test accuracy of $91.50\%$ on the Fashion-MNIST dataset \cite{fashion-mnist}. We design a variety of attacks using the \textit{cleverhans} toolbox \cite{papernot2016cleverhans}, and test our defense on this classifier for all the following experiments. The proposed defense involves the projection operation, following algorithm \ref{mainAlg}, where the unknown corruptions are adversarial perturbations. The performance of different defense strategies is measured using $500$ randomly chosen test images from the dataset, which are cleaned in batches of size $100$. We use the following hyper-parameter settings $T_1 = 10, T_2 = 10,\gamma_s = 1e-2, \gamma_g = 8e-2$ that are determined using a validation set of 100 examples. We observe that, as before, these settings are not very sensitive, and remain effective over a large range of values. 

\paragraph{\textbf{Universal perturbations:}} \suttl~ also provides effective defense against universal perturbations \cite{moosavi2017universal,bbuni}, which belong to the class of image-agnostic perturbations where an attack is just a single vector which when added to the entire dataset can fool a classifier. To test this defense, we first design a targeted universal perturbation using the Fast Gradient Sign Method (FGSM)~\cite{GoodfellowSS14}, by computing the mean adversarial perturbation from $N=15$ test images, i.e. for an adversarially perturbed image $\tilde{X}$, we define the universal perturbation to be $\nu_{u} = \sum_i(X_i-\tilde{X})/N$. We can also increase the magnitude of the attack by scaling it: $\tilde{X}_{u} = X + \alpha \nu_{u}$. Typically, a larger magnitude implies a stronger attack up to a certain point after which it becomes noise and reduces to a trivial attack. In Figure \ref{fig:adv}(a), we observe that when compared to the state-of-the-art \textit{Defense-GAN}, our defense is significantly more robust.

\paragraph{\textbf{Image-dependent attacks:}} We test MimicGAN's defense against the following image-specific attacks: (a) The Carlini-Wagner L2 attack (CWL) \cite{CWL2}, (b) Fast Gradient Sign Method (FGSM) \cite{GoodfellowSS14}, (c) Projected Gradient Descent Method (PGDM) \cite{madry2018towards}, (d) DeepFool \cite{moosavi2016deepfool}, (e) Basic Iterative Method (BIM) \cite{kurakin2016adversarial} and (f) Obfuscated Gradients \cite{athalye2018obfuscated}. We hypothesize that even though the perturbation on each image is different, the surrogate learns an \emph{average} perturbation when presented with a few adversarial examples. As seen in Table \ref{fig:adv}(b), this turns out to be a strong regularization, resulting in a significantly improved defense compared to baseline approaches such as \textit{Defense-GAN} \cite{defenseGAN} or \textit{Cowboy} \cite{santhanam2018defending}. {\color{blue} We use the same GAN backend for these baselines so that the only variable in the adversarial defence is the projection technique. The defence performance reported in Figure \ref{fig:adv}(b) is obtained using 3 separate GAN backends that are trained with random subsets containing 45000 images or $\sim80\%$ of the MNIST training set, to ensure they result in sufficiently different GANs}. The obfuscated attack \cite{athalye2018obfuscated} is the strongest attack considered here as it targets the GAN in addition to the classifier, we attack all the three GANs and report the performance. While \suttl~is vulnerable to such an attack, it can afford a stronger defense than plain GAN-based defenses, as seen in Table \ref{fig:adv}(b).

\subsection{\textbf{Anomaly Detection}}

GANs have become a popular choice for unsupervised anomaly detection since out of distribution samples are represented as those samples with a relatively high reprojection error, since they are not well represented by the image manifold inferred by the generator. A typical experimental setup \cite{zenati2018efficient,akcay2018ganomaly,an2015variational} for a $k$ class dataset is to train a GAN on $k-1$ classes, and use the $k^{th}$ class as the anomaly - this is repeated for every class. We test the effectiveness of \suttl~ in such a task, which we expect to improve with more robust projections. We also compare our method with recent manifold-projection based anomaly detection techniques \cite{an2015variational,akcay2018ganomaly,zenati2018efficient}. 

\paragraph{\textbf{Experimental Details:}} We perform the anomaly detection task on the MNIST digits training set, where we leave one class out and train a GAN on the remaining classes. We use the same hyper-parameters used in the domain adaptation setting described in section \ref{sec:da}. The training set contains $80\%$ of the normal data, and the test set contains the remaining $20\%$ of the normal, along with all the anomalous class samples, similar to \cite{zenati2018efficient}. We use the final projection error given in eq. (\ref{eq:proj_error}) to distinguish normal samples from anomalous ones, the hypothesis being that a normal sample should have a significantly lower error than one that is out of distribution. Though some recent approaches have also used distance in the feature space of the discriminator as a measure for detection, we found the projection error to be more effective. We compute the area under the ROC curve as the evaluation metric for detection performance. 

\begin{figure}[!htb]
	\centering
\includegraphics[trim={0.0cm 1cm 2.0cm 0.0cm},clip,width=0.95\linewidth]{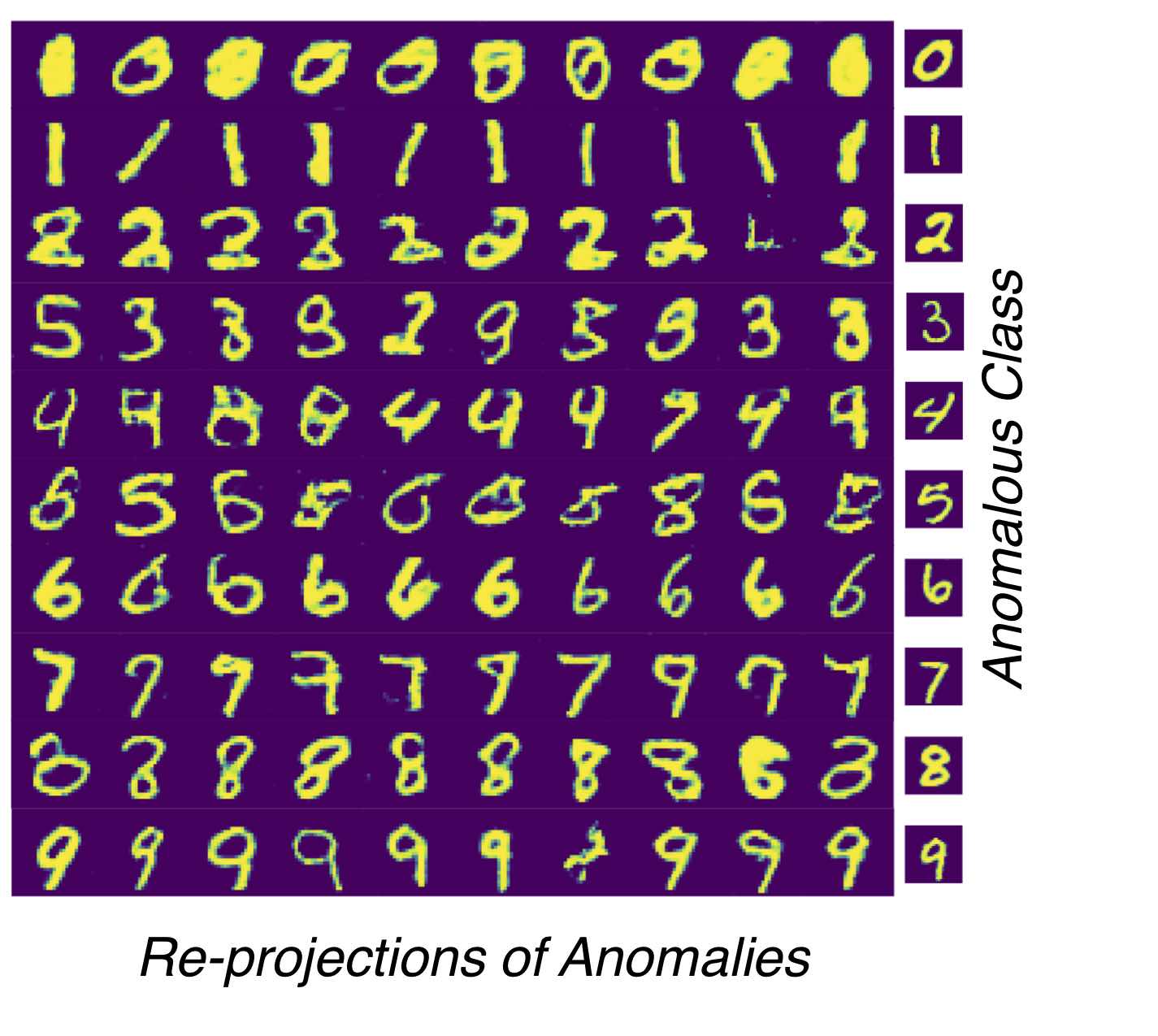}\label{fig:anomaly}
\caption{\suttl~ for anomaly detection using a leave one class out setup. Here, the class indicated as the anomaly is left out during training a GAN. Next, the anomalous class is reprojected onto this GAN. We expect a better projection technique to identify out-of-distribution samples more effectively.}
\end{figure}
Table \ref{tab:anoGAN} shows the average detection performance across all $10$ classes in the MNIST dataset, in comparison to several recent techniques.  We see that using corruption-mimicking with \suttl~ significantly improves detection performance. In addition, we match the performance of a state-of-the-art technique specifically designed for anomaly detection, simply through robust reprojection. It is also worth noting that over our baseline approach, PGD, \suttl~ increases detection performance by nearly 30 percentage points. 

\begin{table}[t]
\centering

	\begin{tabular}{p{1.6in}p{0.8in}}
		\hline
	Method & \small{Area under ROC}\\\hline
	VAE\cite{an2015variational} & $0.34\pm 0.18$ \\
	AnoGAN\cite{schlegl2017unsupervised} & $0.44 \pm 0.07$ \\
	GANomaly (BIGAN) \cite{akcay2018ganomaly} & $\mathbf{0.78} \pm 0.11$  \\
	EGBAD (PGD baseline) \cite{zenati2018efficient} & $0.50 \pm 0.13$\\
	MimicGAN (\emph{ours}) &$\mathbf{0.78}\pm 0.14$\\
	\hline
\end{tabular}
\caption{{Anomaly Detection on MNIST leave-one-class out experiment. Average performance on all 10 classes is reported below. Corruption mimicking boosts PGD by $30\%$ to match state of the art techniques.}}
\label{tab:anoGAN}
\end{table}

\section{Discussions and Future Work}
In this paper, we presented \suttl, an entirely unsupervised system that can accurately project images back onto the image manifold even in presence of a variety of \emph{unknown} corruptions. We achieve this by introducing a corruption mimicking surrogate network, in addition to finding a solution in the latent space of a generative model without any additional data augmentation or supervision. The properties of the surrogate network enables robustness to a variety of corruptions, for example we show that a surrogate with a spatial transformer layer provides robustness to affine transformations. We show that by improving the robustness of projections across these corruptions, a huge boost in performance can be obtained in a wide variety of applications leveraging generative priors such as: adversarial defense, domain adaptation, and anomaly detection. The results in this study indicate that such learned priors can be much more powerful than previously understood when coupled with robust projection strategies.
\vspace{5pt}

\noindent There are several avenues of future study, particularly with respect to the surrogate network which is central to \suttl. For applications like domain adaptation, including available supervision in the source domain may lead to improved task-specific alignment, thereby improving adaptation quality. Next, since the surrogate network is trained at test time with a few observations, it maybe easy to break it using few adversarial examples, that are crafted with knowledge of the surrogate network. The current framework cannot handle such shifts and will require generilzations that can provide reliable projections even under adversally perturbed observations. Exploring how robust projections improve problems in inverse imaging remains to be addressed. In this regard, a variational projection can also be useful in recovering multiple plausible solutions to under-determined inverse problems.

\subsection*{\textbf{Disclaimer}}
{\small
 \noindent This document was prepared as an account of work sponsored by an agency of the United States government. Neither the United States government nor Lawrence Livermore National Security, LLC, nor any of their employees makes any warranty, expressed or implied, or assumes any legal liability or responsibility for the accuracy, completeness, or usefulness of any information, apparatus, product, or process disclosed, or represents that its use would not infringe privately owned rights. Reference herein to any specific commercial product, process, or service by trade name, trademark, manufacturer, or otherwise does not necessarily constitute or imply its endorsement, recommendation, or favoring by the United States government or Lawrence Livermore National Security, LLC. The views and opinions of authors expressed herein do not necessarily state or reflect those of the United States government or Lawrence Livermore National Security, LLC, and shall not be used for advertising or product endorsement purposes. }


\bibliographystyle{ieee}      
\bibliography{refs}

\end{document}